\documentclass[10pt,twocolumn,letterpaper]{article}

\usepackage{cvpr}
\usepackage{times}
\usepackage{epsfig}
\usepackage{graphicx}
\usepackage{amsmath}
\usepackage{amssymb}
\usepackage{array}
\usepackage{booktabs}
\usepackage{afterpage}
\usepackage{subcaption}
\usepackage{algorithm}
\usepackage{algpseudocode}
\usepackage{algorithmicx}
\usepackage{authblk}

\usepackage[breaklinks=true,bookmarks=false]{hyperref}

\cvprfinalcopy 

\def\cvprPaperID{****} 

\begin{document}

\title{Robust Optical Flow Estimation in Rainy Scenes}
\author[1]{Li Ruoteng}
\author[1,2]{Robby T. Tan}
\author[1]{Loong-Fah Cheong}
\affil[1]{National University of Singapore} \affil[2]{Yale-NUS}
\maketitle

\begin{abstract}
Optical flow estimation in the rainy scenes is challenging due to background degradation introduced by rain streaks and rain accumulation effects in the scene. Rain accumulation effect refers to poor visibility of remote objects due to the intense rainfall. Most existing optical flow methods are erroneous when applied to rain sequences because the conventional brightness constancy constraint (BCC) and gradient constancy constraint (GCC) generally break down in this situation. Based on the observation that the RGB color channels receive raindrop radiance equally, we introduce a residue channel as a new data constraint to reduce the effect of rain streaks. To handle rain accumulation, our method decomposes the image into a piecewise-smooth background layer and a  high-frequency detail layer. It also enforces the BCC on the background layer only. Results on both synthetic dataset and real images show that our algorithm outperforms existing methods on different types of rain sequences. To our knowledge, this is the first optical flow method specifically dealing with rain.
\end{abstract}

\afterpage{ %
\begin{figure}
\begin{tabular}[t]{ c c }
    \begin{subfigure}{0.225\textwidth}
        \includegraphics[width=1.0\linewidth]{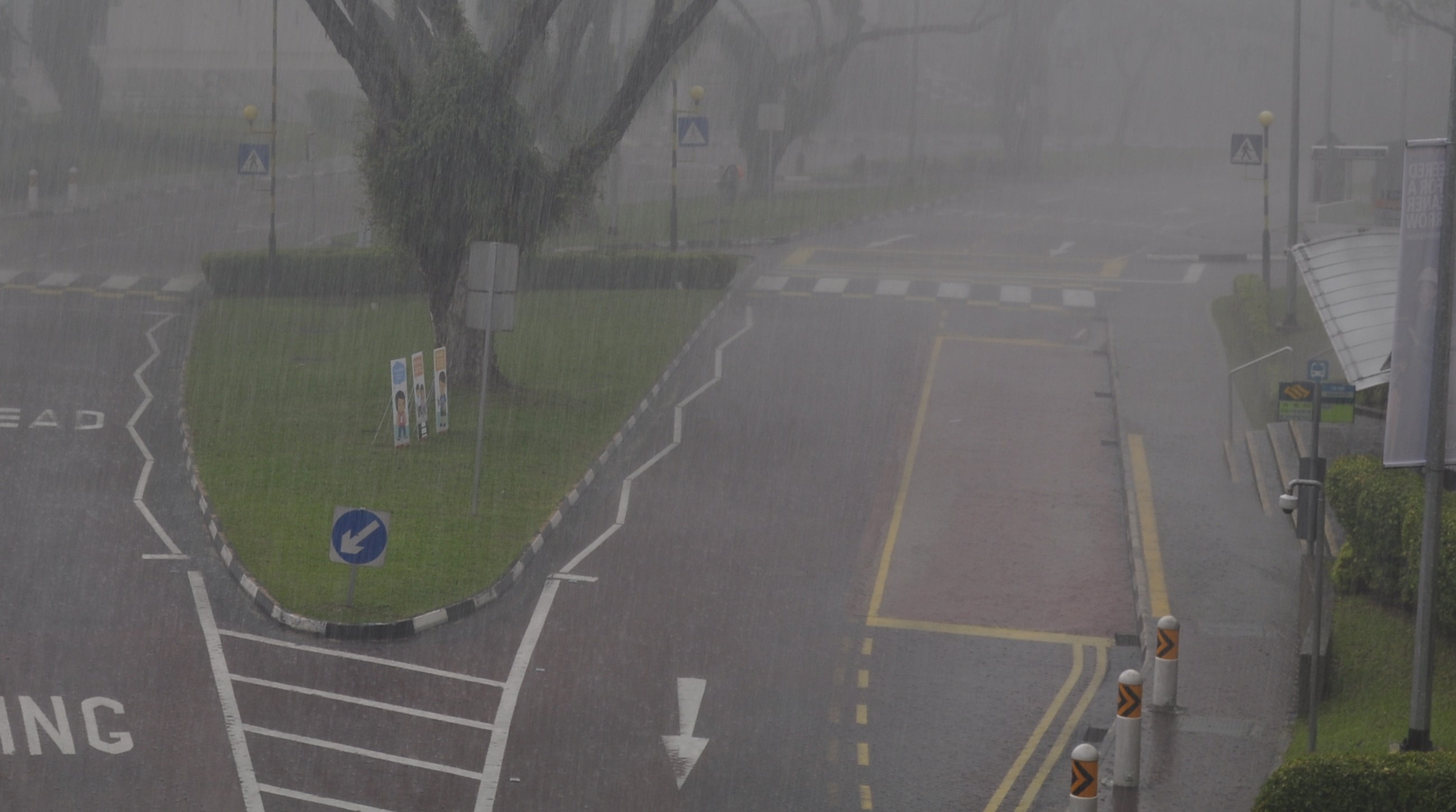}
        \caption{First frame}
    \end{subfigure} &
    \begin{subfigure}{0.225\textwidth}
        \includegraphics[width=1.0\linewidth]{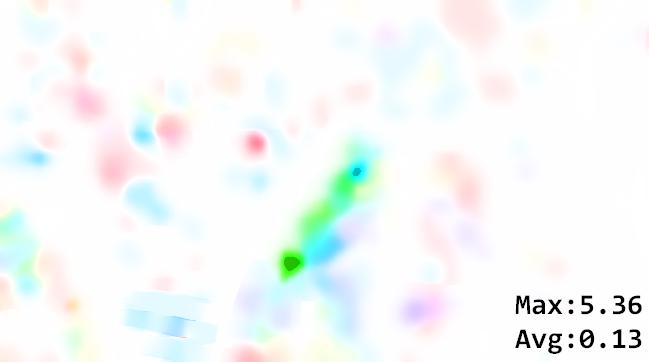}
        \caption{SPMBP \cite{Yuli2015SPMBP}.}
    \end{subfigure}
\\
    \begin{subfigure}{0.225\textwidth}
        \includegraphics[width=1.0\linewidth]{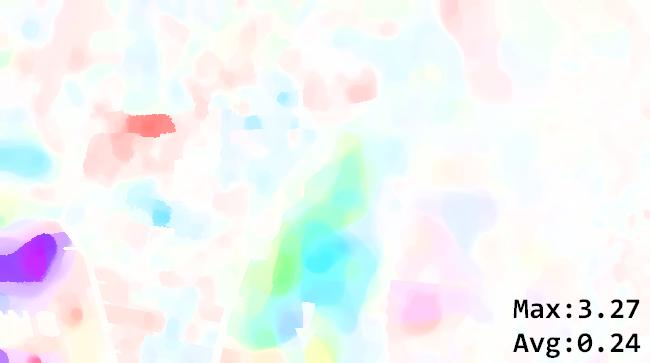}
        \caption{Classic+NL \cite{Li_2016_CVPR}}
    \end{subfigure} &
    \begin{subfigure}{0.225\textwidth}
        \includegraphics[width=1.0\linewidth]{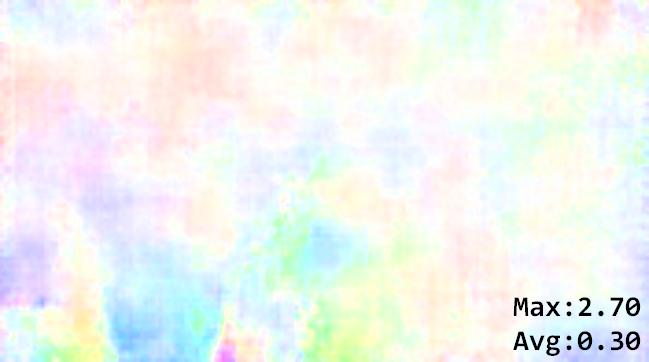}
        \caption{FlowNetS \cite{DFIB15}}
    \end{subfigure}
\end{tabular}
\caption{Comparison of methods on two consecutive frames of a rainy scene with static background\protect\footnotemark. Every object except for the rain in the scene is static, so genuine optical flow should be zero (represented by white) everywhere. The intensity of the color plots thus indicates the magnitude of the errors. The maximum flow amplitude and average flow amplitude (in pixel) of each method are denoted at the corner of each flow field. The flow field visualization of our algorithm is \textbf{perceptually white} (see also Fig~\ref{fig:StaticAnalysis}). (Maximum flow: \textbf{0.0018} px. Average flow: \textbf{0.000195} px)}
\label{fig:CoverPage}
\end{figure}
\footnotetext{For comparison purpose, the flow is normalized to the same scale for visualization for all the methods.}
}
\section{Introduction}
Optical flow methods have been developed for many decades, and achieved significant results in terms of accuracy and robustness. They are shown to generally work when applied to outdoor scenes in clear daylight. However, under realistic outdoor conditions, a range of dynamic weather phenomena such as rain, snow, and sleet will pose a grim problem for these methods. In particular, of all the environmental degradations, \cite{Ritcher2017} showed that rain has the most marked detrimental impact on performance. To our knowledge, no methods have been proposed to handle optical flow estimation under rainy scenes. We consider addressing this problem is important, since more and more vision systems, such as self-driving cars and surveillance, are deployed in outdoor scenes, and rain is an inevitable natural phenomena or even an everyday occurrence in some regions of the world. In this paper, we develop an algorithm that can handle rain in optical flow estimation. In the following, our discussion focuses on rain, though the discussion and the resulting formulation are generally applicable to other dynamic weather conditions such as snow and sleet. (Some examples and experiments on snow and sleet can be found in our supplementary material).

The challenge of estimating optical flow in rainy scenes can be categorized into two problems. One problem refers to rain streaks, which due to their dynamic nature, appear in different locations from frame to frame, thus causing violation to the brightness constancy constraint (BCC). The spurious gradients created by the rain streaks also pose problems for the gradient constancy constraint (GCC). The other problem refers to the rain streak accumulation (Fig.~\ref{fig:CoverPage}). Visually rain streaks throughout some space are accumulated and we can no longer see the streaks individually (visually similar to fog). Images affected by rain accumulation generally suffer from a veiling effect and low contrast. Under torrential downpour or heavy snow, the second problem is severe enough to warrant a special mechanism to come to grips with the issue. Existing optical flow methods usually do not consider these degradation effects, and hence fail to produce satisfactory results.

Most existing optical flow methods rely on the brightness and gradient constancy constraints, which in rainy scenes do not hold anymore or become highly susceptible to noise due to the two aforementioned problems. Various robust statistics measures proposed by the community \cite{Black:1996:REM}\cite{Barron:1994:POF:181447.181452}\cite{Wedel:2009:IAT}\cite{Sun:CVPR:10} help treat limited noise from rains, but do not work robustly on image degradation as strong and complex as heavy rains. A direct solution is to apply a deraining method before optical flow computation. However, most of the video-based deraining methods are designed only for rain streaks removal, and assume static background. Crucially for the optical flow estimation problem, most of these \textbf{single image based} deraining methods process each frame independently, and therefore consistency across frames cannot be guaranteed. Moreover, most of the deraining methods introduce artifacts, such as blur around the rain streak region, high frequency texture loss, image color change, etc. These artifacts are also inconsistent in their appearance throughout an image sequence, thus rendering the brightness constancy constraint invalid.

In this paper, our goal is to develop an optical flow algorithm that can work robustly and accurately in the presence of rain streaks and rain accumulation. To achieve this goal, our idea is based on the observation that the radiance of most raindrops has the same intensity on each RGB channel. Hence, by subtracting the maximum channel by the minimum channel of the rain image, we can reduce the rain streak influence in the resultant map (which we call the residue channel).  Besides,  to handle rain accumulation, we use image decomposition to separate the image into a piecewise-smooth background layer which captures the diagnostic structure of the image, and a high-frequency detail layer which contains the noise, rain, and the fine local details of the background.

Our contributions in this paper are (1) proposed residue channel for reducing the effect of rain streaks; (2) a layer decomposition scheme to extract the principal structure of the image, with the latter providing more reliable information under low contrast, (3) a new real rain optical flow dataset for evaluation.

\section{Related Work}
Optical flow algorithms that are robust to noise and outliers have been studied for a long time \cite{Black:1996:REM}\cite{Barron:1994:POF:181447.181452}\cite{Wedel:2009:IAT}\cite{Sun:CVPR:10}. While these techniques may be able to handle a moderate amount of corruptions such as those brought about by a drizzle, they are unlikely to prevail against the heavy corruptions caused by a torrential downpour. Brox et al.'s \cite{Bro04a} utilizes the GCC to improve robustness against illumination change. However, in rainy scenes, rain streaks create spurious gradients which violate the GCC. Compounding these issues is the loss of contrast caused by rain accumulation; it renders both the brightness constancy constraint and gradient constancy constraint highly susceptible to noise.

One of the popular practices in optical flow estimation is to perform some kind of layer separation. Trobin et al.'s \cite{DBLP:conf/dagm/TrobinPCB08} is the first work to introduce structure-texture decomposition denoising \cite{Rudin:1992:NTV:142269.142312} into the computation of optical flow. The purpose is to remove shadow and shading from the texture layer. However, for rainy scenes, high frequency rain streaks will appear in the texture layer and compromise the utility of the texture layer for flow estimation.
Recently Yang et al.'s \cite{jl} proposes a double-layer decomposition framework to handle transparency or reflection, based on the assumption that both layers obey sparse image gradient distributions. This method cannot be used to remove the rain layer since the rain streaks result in a lot of gradients.

Mileva et.al's \cite{Mileva2007} proposes an illumination-robust variational method using color space transformation to handle shadow and highlights. Unfortunately, the HSV colour space and \(r\phi\theta\) color space approaches do not result in measures that are invariant under the effects of rain streaks and hence cannot be directly applied to rainy scenes.

It is beyond the scope of this paper to offer a comprehensive review of the immense optical flow literature, but the emerging deep learning approach certainly deserves a mention. Several Convolutional Neural Network (CNN) approaches \cite{DFIB15}\cite{choy_nips16}\cite{FLowNet2}\cite{DBLP:journals/corr/RanjanB16} demonstrate the possibility of using a deep learning framework to estimate flow, but these methods are meant for optical flow estimation under normal scenes. CNN-based methods are heavily optimized over a lot of training data. Unfortunately, obtaining the optical flow ground-truths for rainy scenes is not easy. This issue is compounded if we want the method to be applicable to not just rain but a variety of dynamic weather phenomena such as snow and sleet. In contrast, our method leverages on the physics of the image formation process; theoretically, it can offer a much more parsimonious solution to a range of problems posed by different weather phenomena.

A number of single-image rain streaks removal methods have been proposed \cite{Kang12Rain}\cite{Li_2016_CVPR}\cite{YangTFLGY16}. Kang et al.'s \cite{Kang12Rain} decomposes an input image into low frequency (rain streak free) and high frequency components, and subsequently extracted geometric details from the high frequency component to recover the de-rained image. Li et al.'s \cite{Li_2016_CVPR}, decomposes the rain image into a rain-free background image layer and a rain streak layer and solves this formulation by introducing GMM priors of the background and rain streaks. Yang et al.'s \cite{YangTFLGY16} incorporates the convolutional neural network to learn the binary rain region features and rain streak intensity features. In the output image of these deraining methods \cite{Kang12Rain}\cite{Li_2016_CVPR}\cite{YangTFLGY16}, the rain streak regions are blurred, and the background geometric details can be lost. Hence, the derained sequences of both approaches violate both the BCC and the GCC. Our experiments show that existing optical flow method with a state-of-the-art deraining pre-processing step does not work properly due to the artifacts introduced by the deraining algorithms.

\begin{figure}[t]
\begin{center}
   \includegraphics[width=0.490\linewidth]{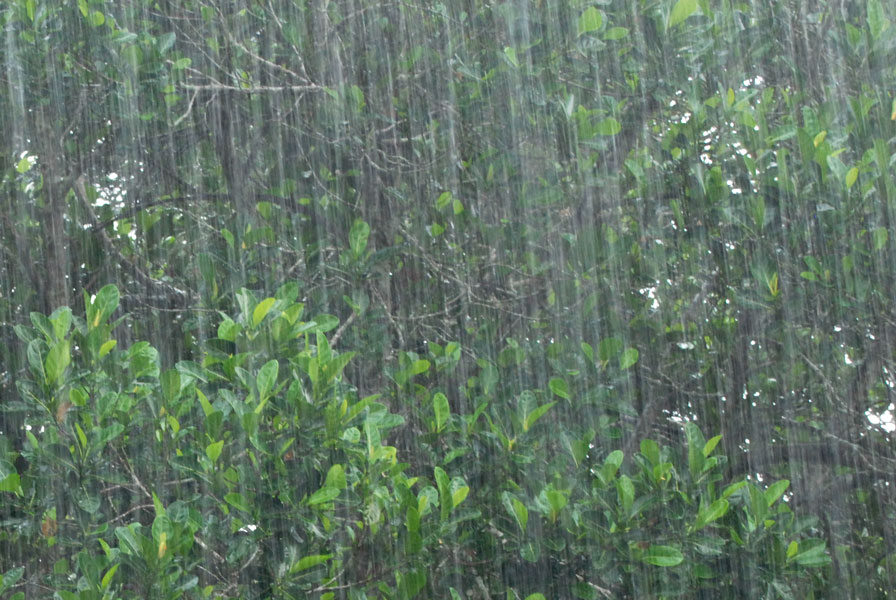}
   \includegraphics[width=0.490\linewidth]{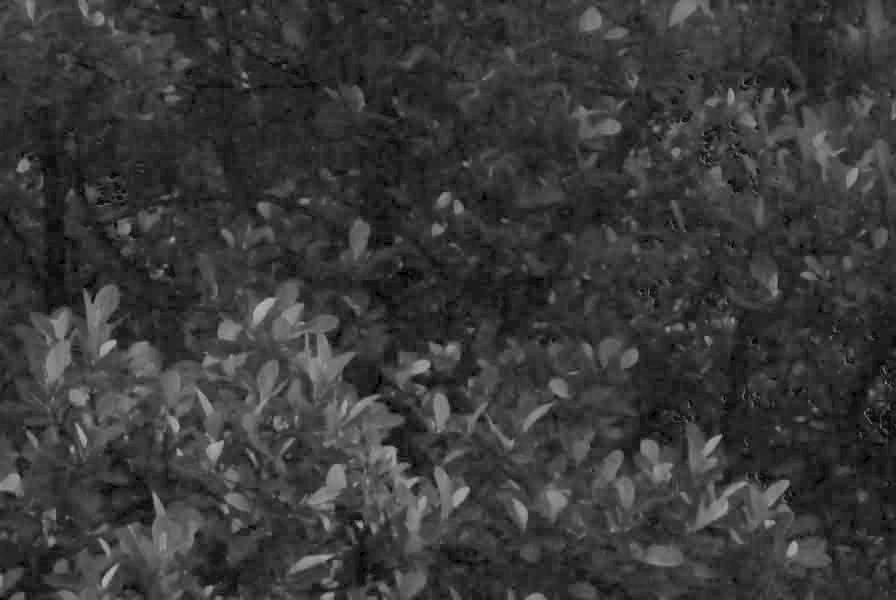}
\end{center}
   \caption{\textbf{Left}: Image captured in a rainy scene with strong rain streaks. \textbf{Right}: Residue channel of this image. The rain streaks are significantly reduced in residue channel.} 
\label{fig:Tree}
\end{figure}

\section{Residue Channel}
The purpose of the residue channel is to reduce the effect of rain streaks based on the observation that the radiance of a raindrop has generally equal intensity in each RGB channel\cite{Zhang:2006:Derain}. The details are as follows. The appearance of rain streaks is caused by the movement of raindrops during the camera exposure\cite{Garg:2007:VR}. If we assume the exposure time is $T$ and the elapsed time while a raindrop is passing through a pixel \(\mathbf{x}\) is \( \tau\), the rain image \( I \) captured by the camera is a linear combination of the average raindrop radiance \( \bar E_r \) and the background radiance \( E_b \):
\begin{equation}
I(\mathbf{x}) = \tau \bar E_r(\mathbf{x}) + (T-\tau) E_b(\mathbf{x}) ,
\label{Eq:1}
\end{equation}
where
\begin{equation*}
    \bar E_r = \frac{1}{\tau} \int\limits_0^\tau E_r \mathbf{d} t , \quad 0 \leqslant \tau \leqslant T .
\end{equation*}
\(E_r\) is the radiance of the raindrop at a particular time. Following \cite{Zhang:2006:Derain}, the radiance of a raindrop on each chromatic channel, \(\bar E_{r}^R, \bar E_{r}^G, \bar E_{r}^B\), is approximately the same:
\begin{equation}
    \bar E_{r}^R = \bar E_{r}^G = \bar E_{r}^B .
\label{Eq:2}
\end{equation}
For a rain image I, we define
\begin{equation}
    I_{res}(\mathbf{x}) = I^{M}(\mathbf{x}) - I^{m}(\mathbf{x}) ,
\label{Eq:3}
\end{equation}
where \( I^M(\mathbf{x})\) and \(I^m(\mathbf{x})\) are the maximum-intensity color channel and minimum-intensity color channel of the rain image \(I\) at pixel \(\mathbf{x}\) respectively. We call \(I_{res} \) the \textit{residue channel} of image \( I \) as shown in Fig.~\ref{fig:Tree}.

By combining Eq.(\ref{Eq:1}) and Eq.(\ref{Eq:2}), the radiance due to the raindrops is cancelled:
\begin{equation*}
        {I_{res}} = (T-\tau)  (E_b^{M} - E_b^{m}).
\end{equation*}
Thus, the residue channel is linearly proportional to the channel difference of the background radiance alone:
\begin{equation}
    I_{res}  = \alpha (E_b^{M} - E_b^{m}) ,
\label{Eq:4}
\end{equation}
where \( \alpha = T - \tau \) and \( 0 \leqslant \alpha \leqslant T \) .

\begin{figure}[t]
\begin{center}
   \includegraphics[width=0.25\linewidth]{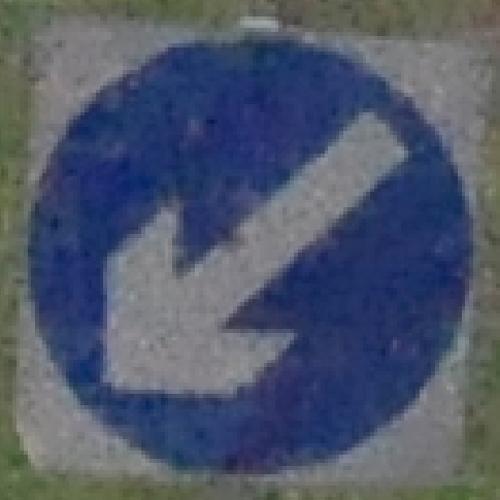}
   \includegraphics[width=0.25\linewidth]{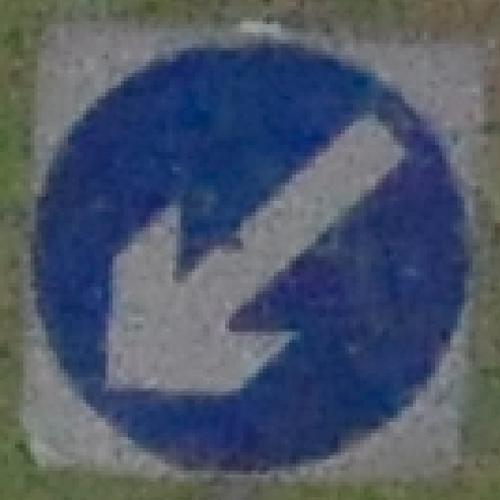}
   \includegraphics[width=0.25\linewidth]{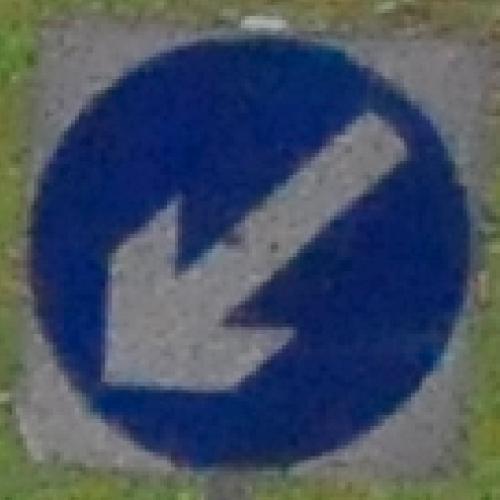}
   \includegraphics[width=0.25\linewidth]{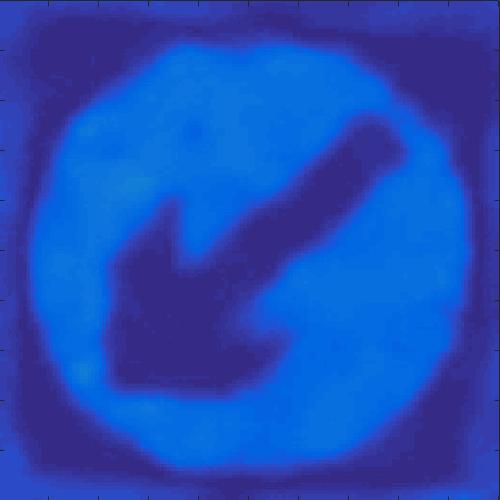}
   \includegraphics[width=0.25\linewidth]{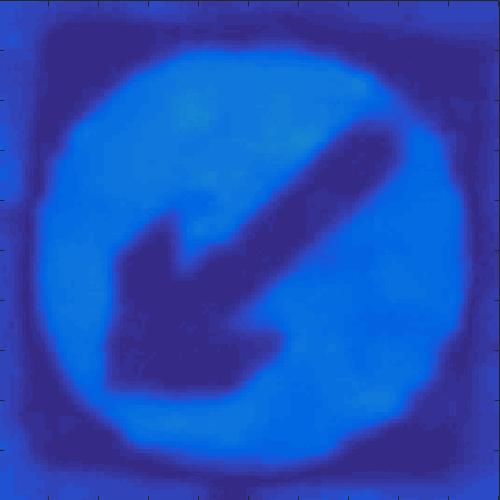}
   \includegraphics[width=0.25\linewidth]{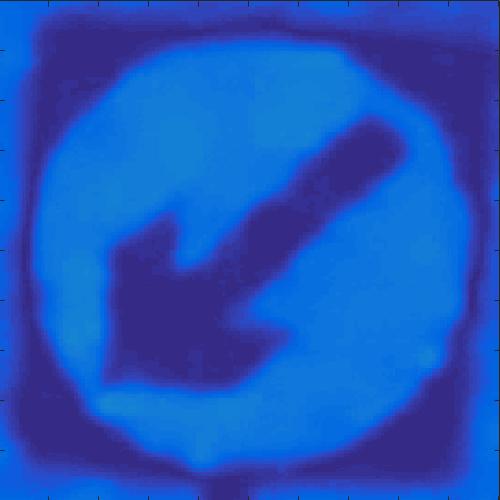}
\end{center}
   \caption{Rain streaks comparison on rainy images (\textbf{Top}) and their residue channels (\textbf{Bottom}). \textbf{Left:} A sign board captured in rain at time t. \textbf{Middle:} The same sign board captured in rain at time t+$\delta$t. \textbf{Right:} Sign board captured in clear daytime. One may find the rain streaks are significantly reduced in the circle area of the sign board. }
\label{fig:Arrow}
\end{figure}

\begin{figure*}[t]
\begin{center}
   \includegraphics[width=1.0\linewidth]{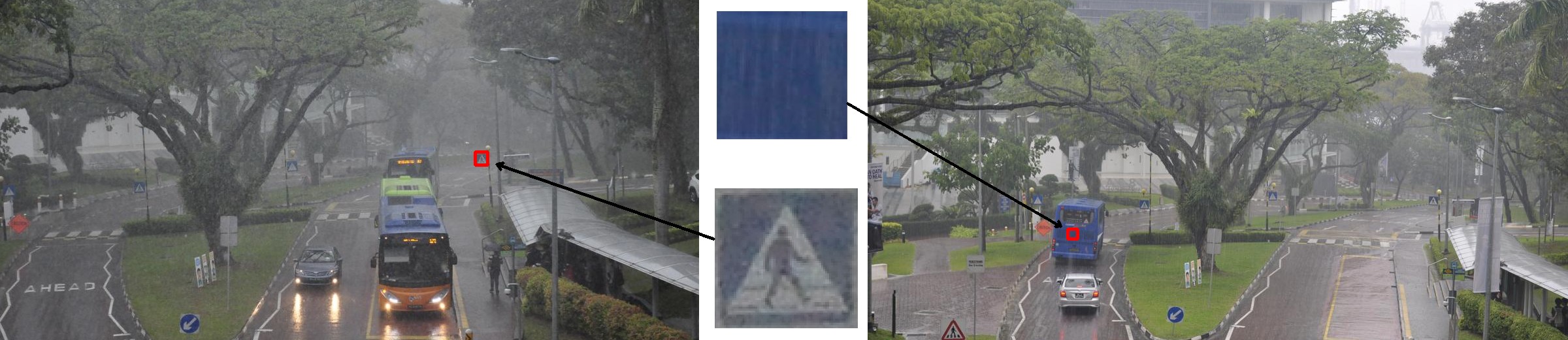}
\end{center}
   \caption{ Rain accumulation (Left) and rain streaks (Right) effect on images. The sign board suffers from poor visibility due to rain accumulation effect. }
\label{fig:RainEffect}
\end{figure*}
We consider the residue channel to be more robust to rain than the original RGB channels for optical flow estimation. Fig.~\ref{fig:Arrow} shows the sporadic noise points introduced by the rain streaks in the residue channel is much smaller than that in the original rain image. We have examined more than 1000 captured rain image to verify this observation (details shown in supplementary material). Based on these observation we can see that the intensity variance caused by rain in the residue channel is considerably smaller than that of the RGB channels. The rain streaks have been significantly reduced in the residue channel. This is a strong support for our residue channel hypothesis.

\begin{figure}[t]
\begin{center}
   \includegraphics[width=0.3\linewidth]{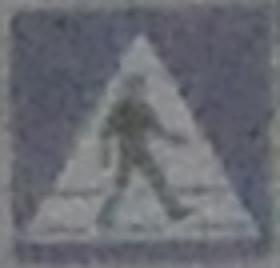}
   \includegraphics[width=0.3\linewidth]{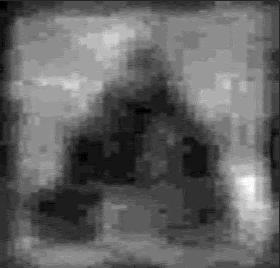}
\end{center}
   \caption{Sign board (\textbf{Left}) and its residue channel (\textbf{Right}) under heavy rain accumulation. The residue channel has been normalized to [0,255] for better visibility.  The contrast of the residue channel is poorer than the rain image. }
\label{fig:Triangle}
\end{figure}

\paragraph{Rain Accumulation} Rain images typically have a severe rain accumulation effect particularly in the heavy rainy scenes (\eg the remote objects in Fig.~\ref{fig:RainEffect}, Left). For each pixel, the intensity contributed by the rain is the accumulation of all the raindrops along the line of sight from the camera to the background object. Thus, we model the pixel intensity by taking the summation of the radiance of the background object and radiance of all the raindrops passing through this pixel during the exposure time.
\begin{equation}
    I(\mathbf{x}) = \sum_i^N \alpha_i \bar E_i(\mathbf{x}) + (1 - \sum_i^N\alpha_i) E_b(\mathbf{x}) ,
\label{Eq:5}
\end{equation}
where N is the number of accumulated rain-streaks along the line of sight with respect to a pixel, $\mathbf{x}$. And \(\alpha_i\) denotes the ratio of the radiance of each raindrop to the entire radiance received by the camera. Because of light scattering and attenuation, the rain image will suffer from low contrast (Fig.~\ref{fig:Triangle} Left), the severity of which depends on the amount of rain accumulation. Generating the residue channel of this image, the contrast turns out to be even lower, since this is based on the residual difference between channels (Fig.~\ref{fig:Triangle} Right). In such low contrast images stemming from severe rain accumulation, the background texture cannot provide reliable matching due to the noise and raindrops. We can only rely on information supplied by the coarse version of the principal regions of the image, including the object boundaries. For this purpose, we embed in our method a decomposition step that separates the image into a piecewise-smooth layer and a fine-details layer.

\section{Proposed Method}
\subsection{Residue Map}

In the variational framework, the optical flow objective function is expressed as:
\begin{equation}  
        E(\mathbf{u}; I_1,I_2) = \Phi_D[I_1(\mathbf{x}) - I_2(\mathbf{x}+\mathbf{u})] + \lambda_s \Phi_S(\nabla \mathbf{u}) ,
\label{Eq:6}
\end{equation}
where \(I_1, I_2 \) are the input sequences with spatial index \(\mathbf{x}\). \(\mathbf{u}\) is the flow vector with \(\lambda_s\) as a regularization parameter and \(\phi_D\) and \(\phi_S \) are the data and spatial penalty functions. We include in our objective function an additional data constraint based on the residue map with a corresponding weighting parameter:
\begin{equation}  
    \begin{split}
     E(\mathbf{u}; I_1,I_2) &= (1-\mathbf{w}) \odot \Phi_D[I_1(\mathbf{x}) - I_2(\mathbf{x}+\mathbf{u})] \\
      & \quad + \mathbf{w} \odot \Phi_D[R_1(\mathbf{x}) - R_2(\mathbf{x}+\mathbf{u})]  \\
      & \quad + \lambda_s \Phi_S(\nabla \mathbf{u}),
    \end{split}
\label{Eq:7}
\end{equation}
where \(R_1,R_2\) are residue channels of rain image \( I_1, I_2 \) respectively. $\odot$ represents element-wise multiplication. \(w\) is the weighting factor defined as follows:
\begin{equation}  
    \mathbf{w} = \gamma \sqrt{(I_1^R - I_1^G)^2 + (I_1^G-I_1^B)^2 + (I_1^B- I_1^R)^2 } ,
\label{Eq:8}
\end{equation}
where \(I_1^R, I_1^G, I_1^B \) are the RGB channels of image \(I_1\) and \(\gamma\) is a scaling parameter.
An object or scene with a low level of color saturation will yield low intensity and low contrast in the residue channel. In the extreme case of white and gray objects, their residue intensity would become black. Since low contrast image will be susceptible to noise, we weigh the additional residue data constraint with $w$ that is given by the Euclidean distance between the pixel color and mid-tone gray in RGB color space.

\subsection{Piecewise-smooth + Fine-detail Decomposition}
When rain is relatively heavy, detailed textures on the background are severely corrupted by the ubiquitous raindrops, and are difficult to recover by the regular ROF decomposition. In this heavily degraded scenario, we resort to a more impoverished and coarse version of the scene to supply the constraint on optical flow. This version of the scene will include the principal contours of the image. For this purpose, we decompose the rain image into a piecewise-smooth layer describing the principal regions of the image and a fine-detail layer containing the background textures, raindrops, and noise. Formally, the observed rain image \(I\) can be modeled as a linear combination of the piecewise-smooth layer \(J\) and the fine-detail layer \(L\):
\begin{equation}   
    I = J + L
\label{Eq:9}
\end{equation}
In many cases of heavy rain scenarios like that in Fig.~\ref{fig:RainEffect}, the background details are seriously contaminated by the rain. The fine-details are unreliable so that our method will rely on the piecewise-smooth structure of the image.
Thus the decomposition can be expressed by the following energy minimization form:
\begin{equation}  
    \min\limits_{J} \parallel I - J \parallel^2 + \parallel \nabla J \parallel_0
\label{Eq:10}
\end{equation}
where \(\nabla = (\partial{x}, \partial{y})^T \).
Hence, taking the optical flow into consideration, we obtain the energy minimization problem as our formulation:
\begin{equation}  
    \begin{split}
    &E(J_1, J_2, \mathbf{u}; I_1, I_2) = \Phi_D[J_1(\mathbf{x}) - J_2(\mathbf{x}+\mathbf{u})]  \\
    &\quad + \lambda_s \Phi_S(\nabla \mathbf{u}) + \alpha (||I_1 - J_1 ||^2 + ||I_2 - J_2||^2) \\
    &\quad + \beta (||\nabla J_1||_0 + ||\nabla J_2||_0)
    \end{split}
\label{Eq:11}
\end{equation}
where \(I_1\) \(I_2\) are two input rain frames. \(J_1, J_2\)  are the piecewise-smooth background layers of the two frames respectively. \(\lambda_s \) is the smoothness parameter for the flow \(\mathbf{u}\). \(\beta\) is the parameter controlling the gradient threshold. The higher the \(\beta\) , the fewer boundaries in the piecewise-smooth background layer.

\subsection{Overall Objective Function}
By introducing the residue channels \( R_1, R_2 \) and its corresponding weight parameter \(w\), our overall objective function is:
\begin{equation}   
    \begin{split}
    &E(J_1, J_2, \mathbf{u}; I_1,I_2) = \lambda_d\{(1-\mathbf{w}) \odot \Phi_D[J_1(\mathbf{x}) - J_2(\mathbf{x} + \mathbf{u})] \\
    &\quad + \mathbf{w} \odot \Phi_D[ R_1(\mathbf{x}) - R_2(\mathbf{x}+ \mathbf{u})]\} + \lambda_s \Phi_S(\nabla \mathbf{u}) \\
    &\quad  + \alpha (||I_1 - J_1 ||^2 + ||I_2 - J_2||^2) \\
    &\quad + \beta (||\nabla J_1||_0 + ||\nabla J_2||_0),
    \end{split}
\label{Eq:12}
\end{equation}
where \(R_1 \) and \(R_2\) represent the residue channel maps of each frame correspondingly. Except for the gradients of the \(J\) layers, all the other terms are in $L2$-norm.


\begin{figure*}[t]
\begin{center}
   \includegraphics[width=0.242\linewidth]{figure/Static/static22-img1.jpg}
   \includegraphics[width=0.242\linewidth]{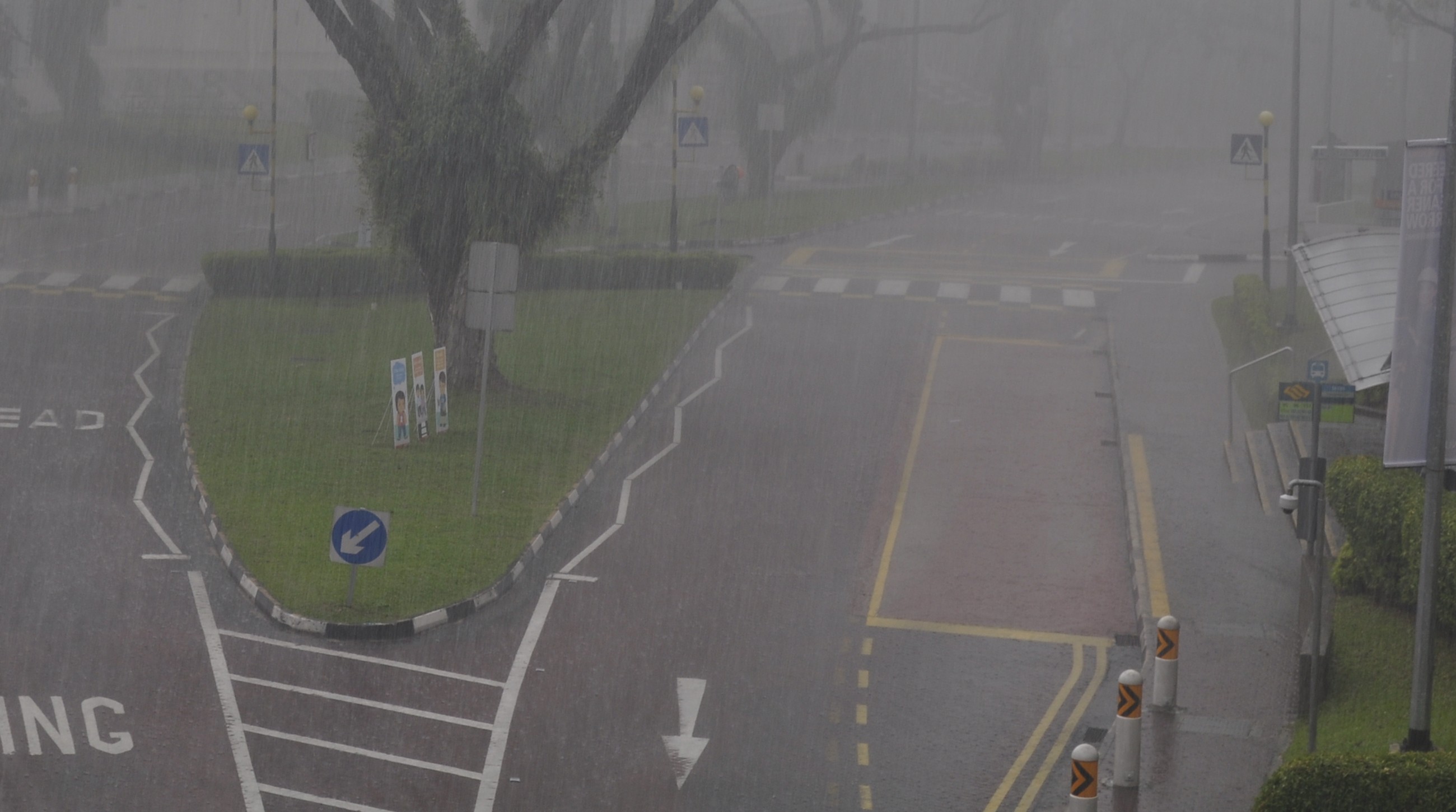}
   \includegraphics[width=0.242\linewidth]{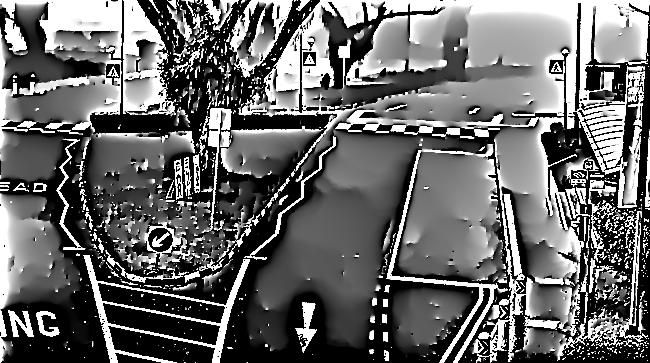}
   \includegraphics[width=0.242\linewidth]{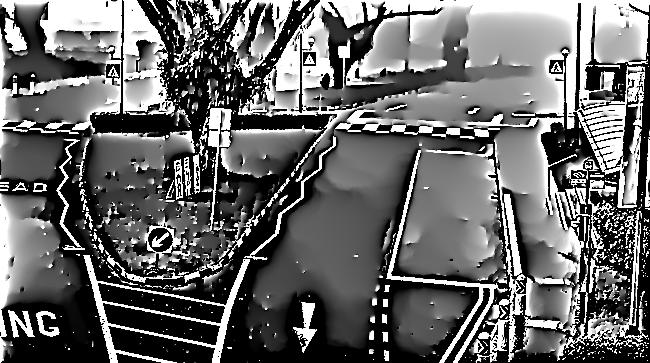}\\
   \includegraphics[width=0.242\linewidth]{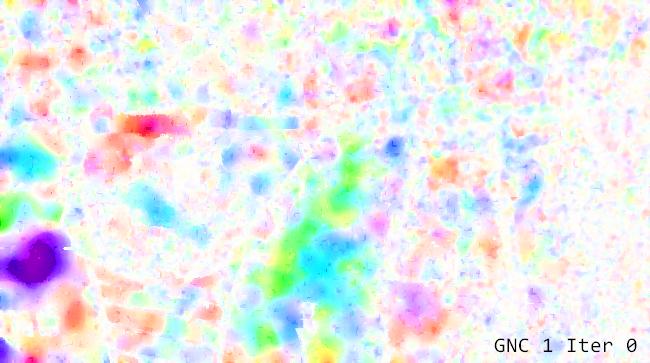}
   \includegraphics[width=0.242\linewidth]{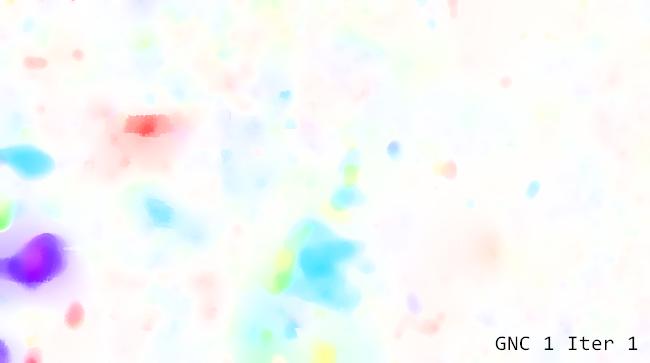}
   \includegraphics[width=0.242\linewidth]{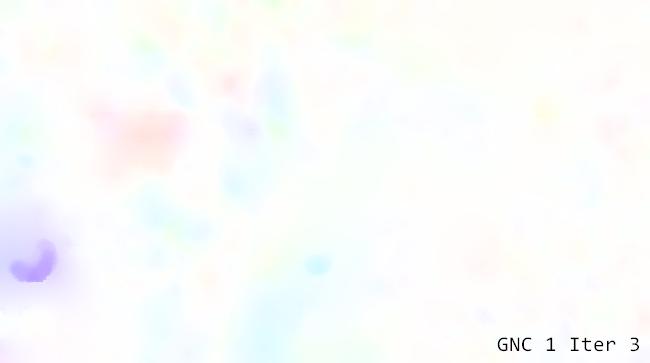}
   \includegraphics[width=0.242\linewidth]{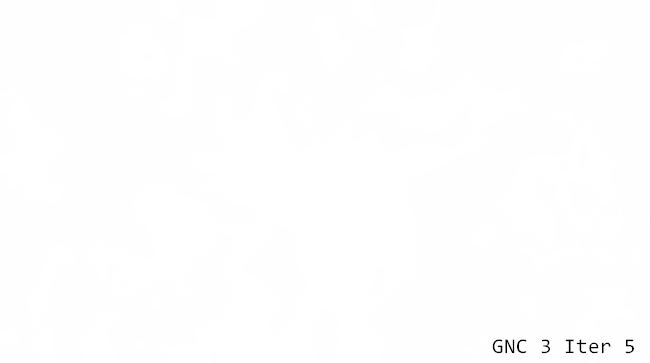} \\
\end{center}
   \caption{\textbf{Top}: Image captured in a rainy scene with strong rain streaks and rain accumulation and the decomposed background layer of this image pair. \textbf{Bottom}: From left to right: the initial estimated flow field before decomposition, intermediate result at GNC level 1, iteration 1 and iteration 2, and the final output flow field at GNC level 3 iteration 5.}
\label{fig:StaticAnalysis}
\end{figure*}


\section{Optimization}

In order to optimize our objective function, we iteratively solve the following sub-tasks given some initialization. \\
\textbf{Sub-problem 1: Optical Flow Computation } Given current piecewise-smooth background layers (\(J_1, J_2\)), we obtain the residue channel maps (\(R_1, R_2\)) and estimate the optical flow vector \(\mathbf{u}\):
\begin{equation}
\begin{split}
\min\limits_{\mathbf{u}} \sum_{\mathbf{x}} & \lambda_d\{(1-\mathbf{w}) \odot \Phi_D [J_1(\mathbf{x}) - J_2(\mathbf{x} + \mathbf{u})] \\
    & + \mathbf{w} \odot \Phi_D[ R_1(\mathbf{x}) - R_2(\mathbf{x}+\mathbf{u})]\} + \lambda_s\Phi_S(\nabla \mathbf{u})
\end{split}
\label{Eq:SubProblem1}
\end{equation}
\textbf{Sub-problem 2: Layer Separation } Given the current optical flow \(\mathbf{u}\), we compute the piecewise-smooth background layer \(J_1\), and \(J_2\) separately:

\begin{equation}   
    \begin{split}
    \min\limits_{J_1} \sum_{\mathbf{x}} \{ & \lambda_d \Phi_D[ J_1(\mathbf{x}) - J_2(\mathbf{x}+\mathbf{u})] + \alpha (||I_1 - J_1 ||^2 \\  &   + \beta (||\nabla J_1||_0  \}
    \end{split}
\label{Eq:Subproblem2a}
\end{equation}

\begin{equation}   
    \begin{split}
    \min\limits_{J_2}\sum_{\mathbf{x}} \{ & \lambda_d \Phi_D[ J_1(\mathbf{x}) - J_2(\mathbf{x}+\mathbf{u})] + \alpha (||I_2 - J_2 ||^2 \\  &   + \beta (||\nabla J_2||_0  \}.
    \end{split}
\label{Eq:SubProblem2b}
\end{equation}
We initialize the optical flow via Eq.(\ref{Eq:SubProblem1}) by assuming the original rain input image as the piecewise-smooth background layer. Eq.(\ref{Eq:SubProblem1}) is solved via standard method in Horn-Schunck based optical flow algorithms. Eq.(\ref{Eq:Subproblem2a}) and Eq.(\ref{Eq:SubProblem2b}) are non-convex because of the $L0$-norm terms. Therefore, we adopt the alternating optimization strategy from \cite{l0smoothing2011}, by introducing two auxiliary variables to decouple the unsmooth gradient term and the smooth quadratic terms. Although there is no guarantee for convergence to this non-convex problem, with initialization as proposed above, this algorithm performs well in practice. In our experiments, we have run our algorithm on hundreds of different rain scenes and it showed good convergence. The details of the steps are summarized in Algorithm 1.

\begin{algorithm}
\caption{ }
\label{alg:pipeline}
\begin{algorithmic}[1]
\State \textbf{Input: }Image sequence \(I_1, I_2\), regularization parameter \(\lambda_s\), parameter \(\alpha, \beta\), maximum iteration M.
\State \textbf{Initialization: } Assign \(J_1^{(0)} \leftarrow I_1, J_2^{(0)} \leftarrow I_2 \). Estimate residue channel \(R_1^{(0)} \leftarrow J_1^{(0)}, R_2^{(0)} \leftarrow J_2^{(0)} \)and initial flow \(\mathbf{u}^0 \leftarrow J_1^{(0)}, J_2^{(0)} , R_1^{(0)}, R_2^{(0)} \)
\For{iteration i = 1, ..., M}
    \State Compute \(J_1^{(i+1)} \leftarrow J_2^{(i)}, I_1, \mathbf{u}^{(i)} \)
    \State Compute \(J_2^{(i+1)} \leftarrow J_1^{(i)}, I_2, \mathbf{u}^{(i)} \)
    \State Compute residue channel \(R_1^{(i+1)}, R_2^{(i+1)}\)
    \State Estimate Flow \item[]
     \(\textbf{u}^{(i+1)} \leftarrow J_1^{(i+1)}, J_2^{(i+1)}, R_1^{(i+1)}, R_2^{(i+1)} \)
\EndFor
\State \textbf{Output: } Estimated flow field \(\mathbf{u}^{(M)}\)
\end{algorithmic}
\end{algorithm}

\section{Experiments}

We evaluate our method by comparing it with representative existing methods \cite{Sun:CVPR:10} \cite{LDOF} \cite{Yuli2015SPMBP} \cite{DFIB15} on synthetic rain (Sect. 6.2) and real rain (Sect. 6.3) data. For synthetic rain, we rendered rain streaks following the rain model from \cite{Garg:2006} on Middlebury \cite{Baker}, Sintel\cite{Butler2012} and KITTI \cite{Menze2015CVPR} optical flow dataset. The rain streaks are generated separately and overlaid on top of the original sequences according to Eq.(\ref{Eq:1}). We render the rain streaks with different intensity and direction randomly. The rain streaks' strength \( \tau \) uniformly varies from 0 to 0.5. The rain streaks direction uniformly varies from -15 degree to 15 degree to the vertical axis. For real rain, we create a new dataset called the Flying Vehicles with Rain (FVR), using a combination of real rain images and synthetic 3D vehicle models applied with affine transformation (for more details, see Sect. 6.1). The camera we use to capture rain images is a NIKON D90 with a focal length of 45mm. All the experiments are run on a desktop with Intel(R) 12-core 3.06 GHz CPU. The time taken to process an image pair with 388x584 image resolution is around 1 minute.

\begin{figure}[t]
\begin{center}
   \includegraphics[width=0.320\linewidth]{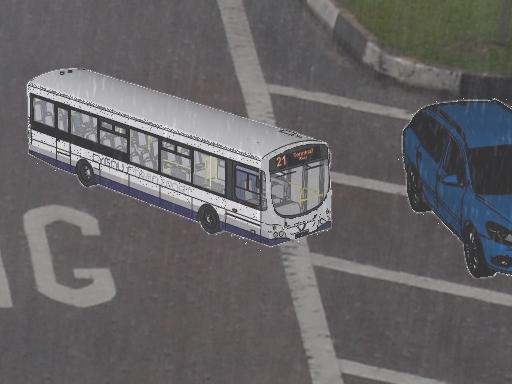}
   \includegraphics[width=0.320\linewidth]{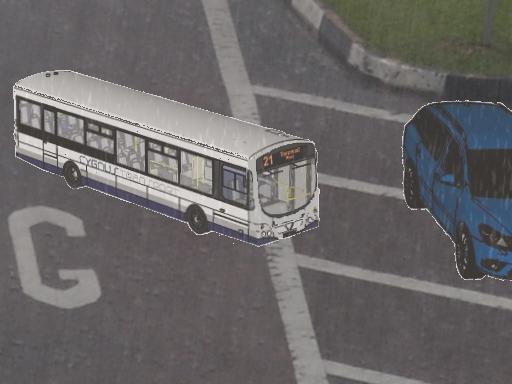}
   \includegraphics[width=0.320\linewidth]{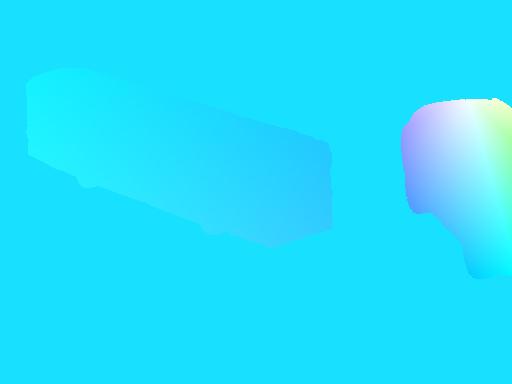} \\
   \includegraphics[width=0.320\linewidth]{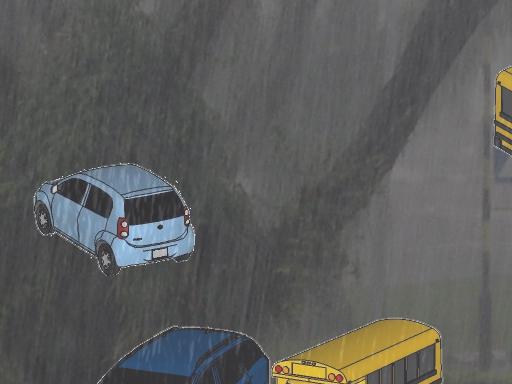}
   \includegraphics[width=0.320\linewidth]{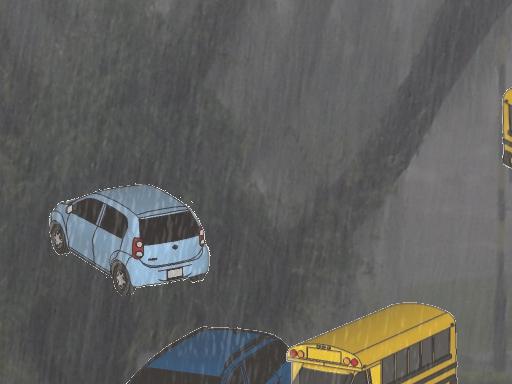}
   \includegraphics[width=0.320\linewidth]{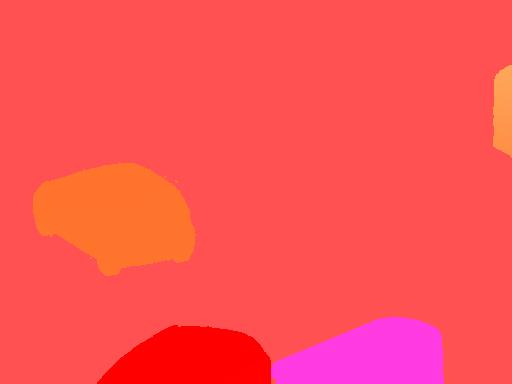} \\
\end{center}
   \caption{Two examples from the Flying Vehicles with Rain (FVR). From left to right are generated image pair and the color coded flow field ground truth.}
\label{fig:FCR}
\end{figure}

\begin{figure*}
\centering
\setlength{\tabcolsep}{1.0pt}
\begin{tabular}{@{} ccccccc @{}}
 & First frame & Classic + NL\cite{Sun:CVPR:10} & SPM-BP \cite{Yuli2015SPMBP} & FlowNetS-rain \cite{DFIB15} & Ours  & Ground truth \\
    R & 	
    \begin{subfigure}{.16\textwidth}
	    \centering
	    \includegraphics[width=1.0\linewidth]{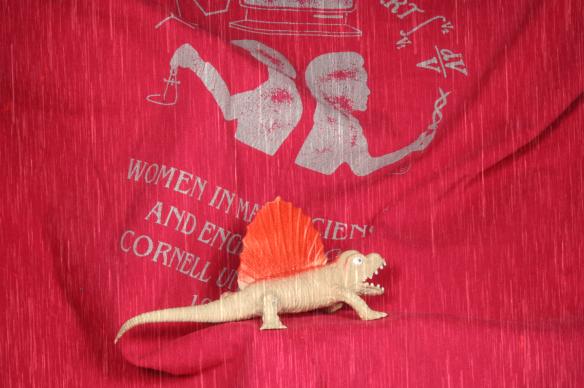}
	    \vspace{-0.35cm}
	\end{subfigure} &
    \begin{subfigure}{.16\textwidth}
	    \centering
	    \includegraphics[width=1.0\linewidth]{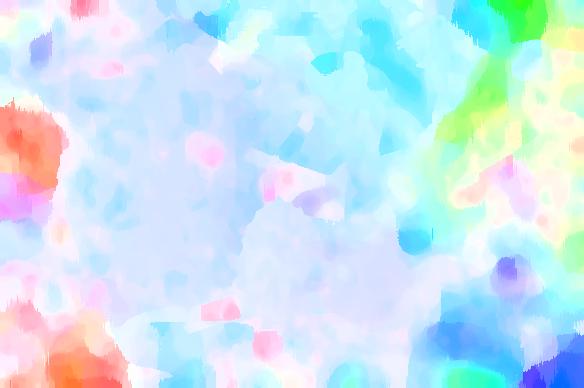}
	    \vspace{-0.35cm}
	\end{subfigure} &
    \begin{subfigure}{.16\textwidth}
	    \centering
	    \includegraphics[width=1.0\linewidth]{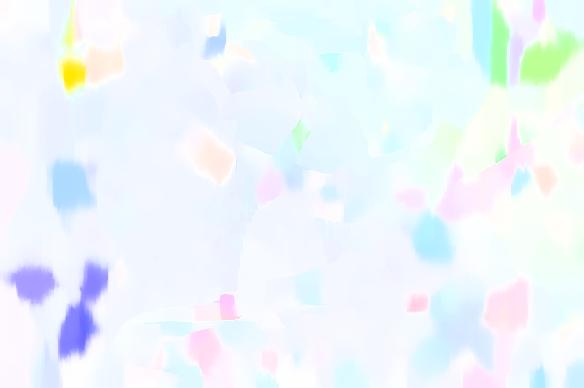}
	    \vspace{-0.35cm}
	\end{subfigure} &
    \begin{subfigure}{.16\textwidth}
	    \centering
	    \includegraphics[width=1.0\linewidth]{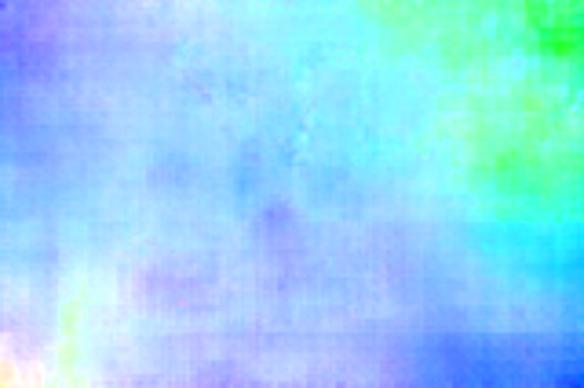}
	    \vspace{-0.35cm}
	\end{subfigure} &
	\begin{subfigure}{.16\textwidth}
	    \centering
	    \includegraphics[width=1.0\linewidth]{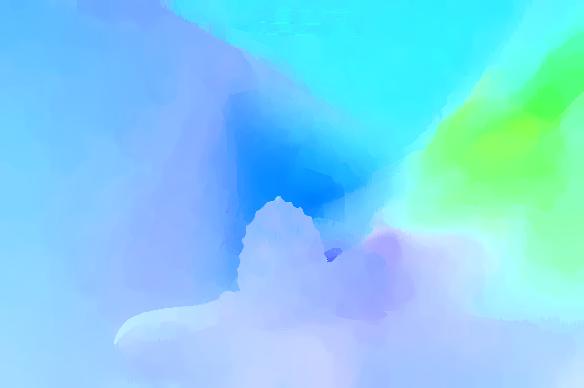}
	    \vspace{-0.35cm}
	\end{subfigure} &
	\begin{subfigure}{.16\textwidth}
	    \centering
	    \includegraphics[width=1.0\linewidth]{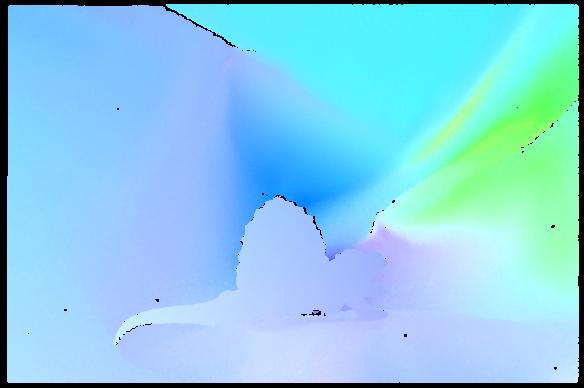}
	    \vspace{-0.35cm}
	\end{subfigure}
\\
    D & 	
    \begin{subfigure}{.16\textwidth}
	    \centering
	    \includegraphics[width=1.0\linewidth]{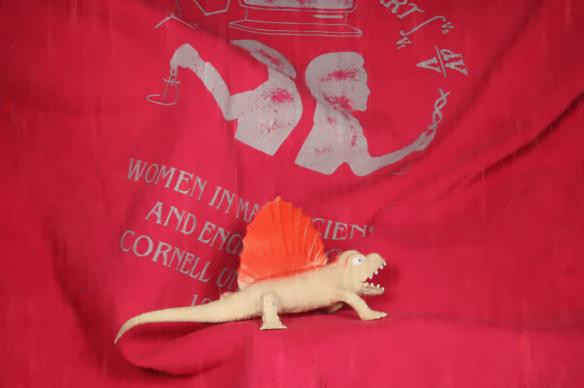}
	    \vspace{-0.35cm}
	\end{subfigure} &
    \begin{subfigure}{.16\textwidth}
	    \centering
	    \includegraphics[width=1.0\linewidth]{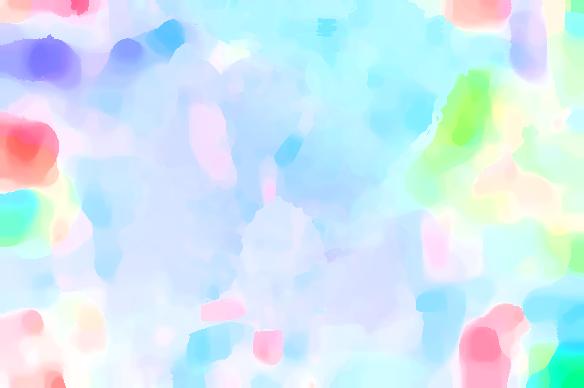}
	    \vspace{-0.35cm}
	\end{subfigure} &
    \begin{subfigure}{.16\textwidth}
	    \centering
	    \includegraphics[width=1.0\linewidth]{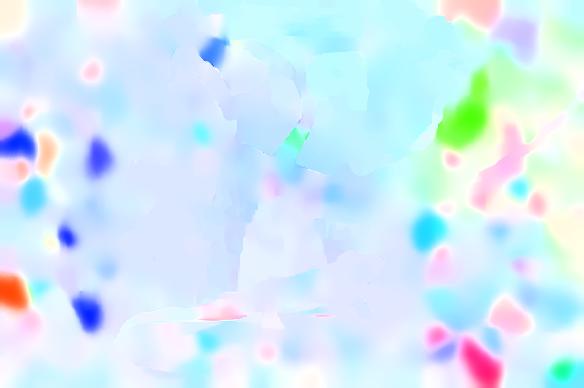}
	    \vspace{-0.35cm}
	\end{subfigure} &
    \begin{subfigure}{.16\textwidth}
	    \centering
	    \includegraphics[width=1.0\linewidth]{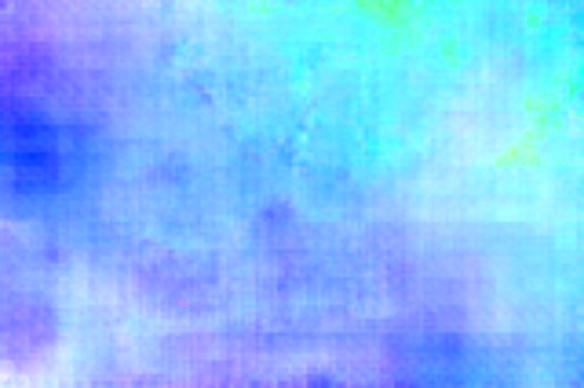}
	    \vspace{-0.35cm}
	\end{subfigure} &
	\begin{subfigure}{.16\textwidth}
	    \centering
	    \includegraphics[width=1.0\linewidth]{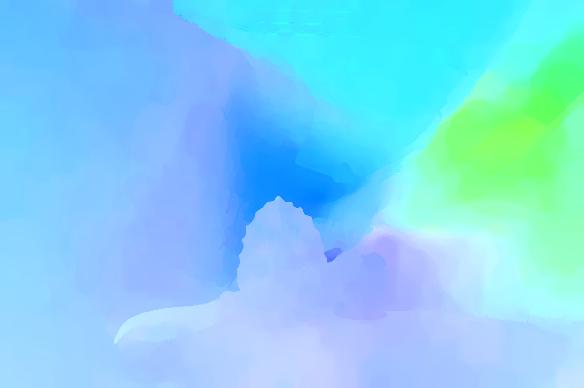}
	    \vspace{-0.35cm}
	\end{subfigure} &
	\begin{subfigure}{.16\textwidth}
	    \centering
	    \includegraphics[width=1.0\linewidth]{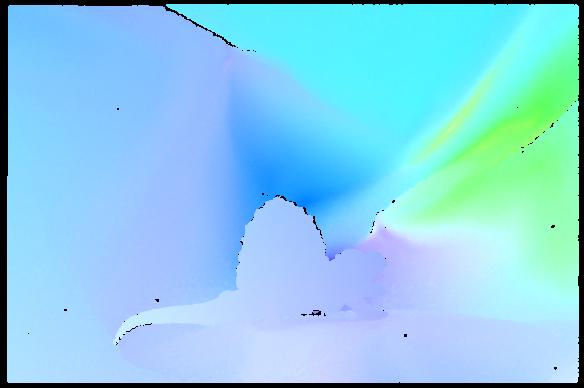}
	    \vspace{-0.35cm}
	\end{subfigure}
\\
   R & 	
    \begin{subfigure}{.16\textwidth}
	    \centering
	    \includegraphics[width=1.0\linewidth]{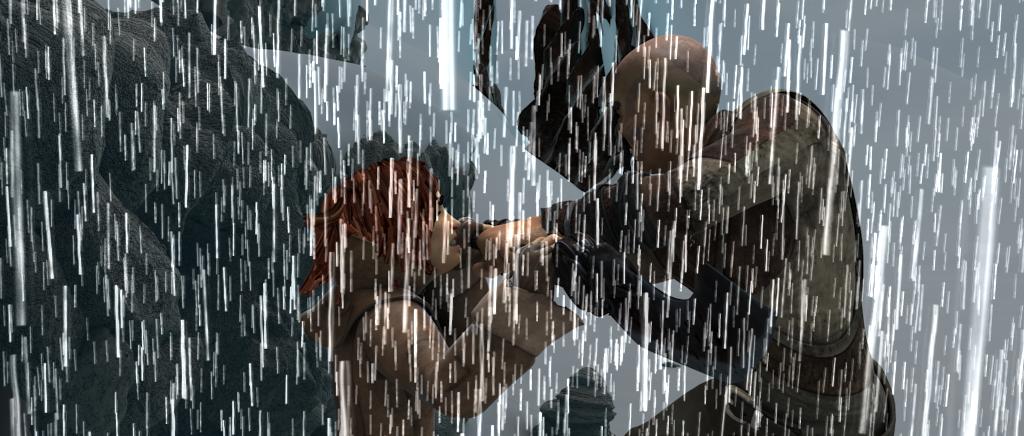}
	    \vspace{-0.35cm}
	\end{subfigure} &
    \begin{subfigure}{.16\textwidth}
	    \centering
	    \includegraphics[width=1.0\linewidth]{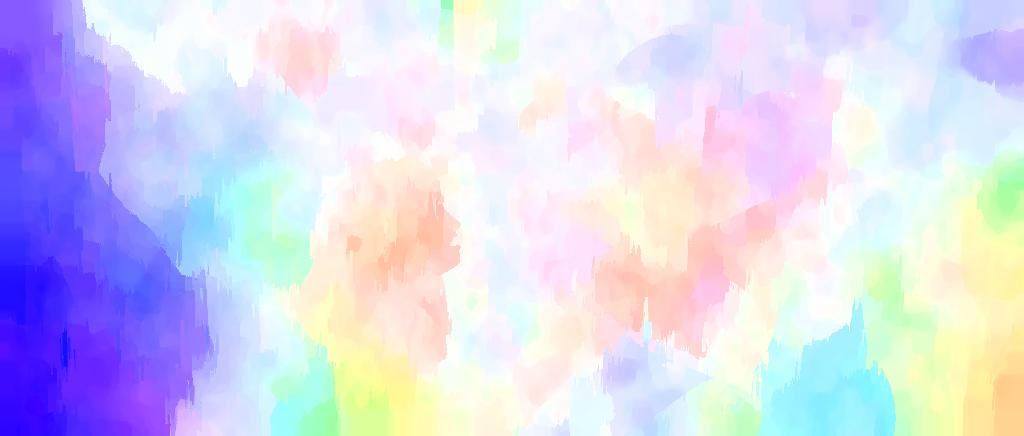}
	    \vspace{-0.35cm}
	\end{subfigure} &
    \begin{subfigure}{.16\textwidth}
	    \centering
	    \includegraphics[width=1.0\linewidth]{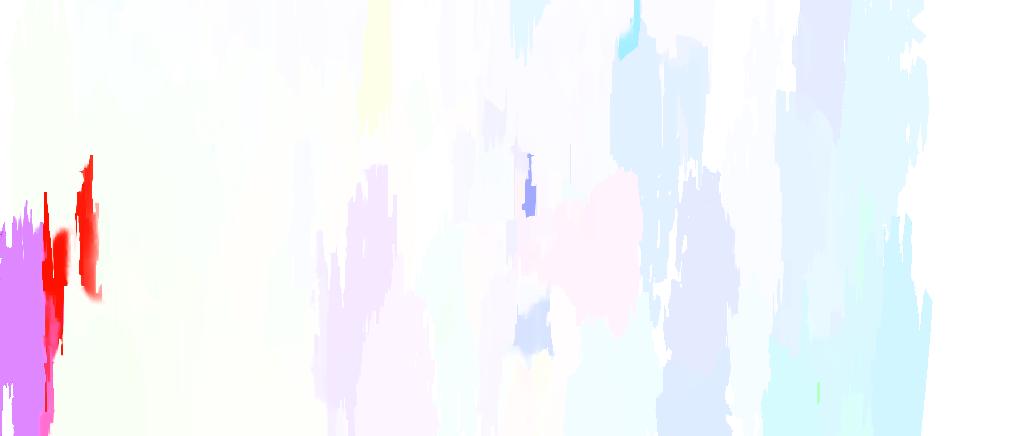}
	    \vspace{-0.35cm}
	\end{subfigure} &
    \begin{subfigure}{.16\textwidth}
	    \centering
	    \includegraphics[width=1.0\linewidth]{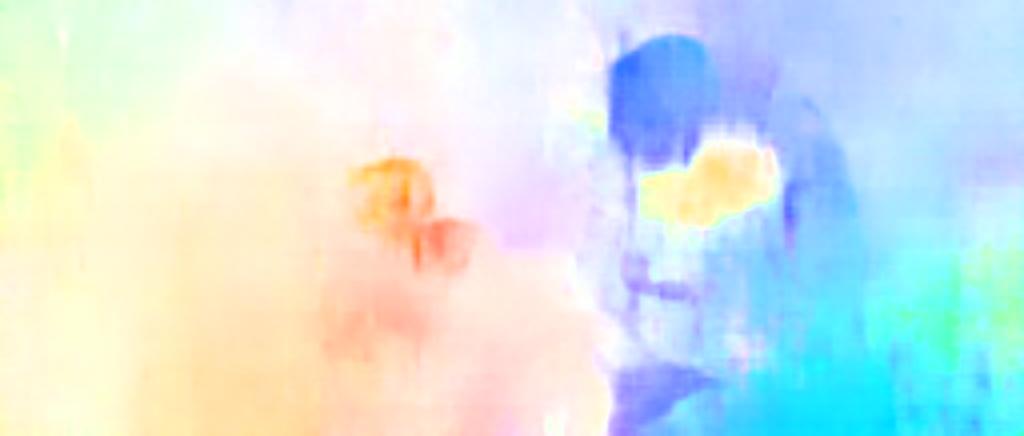}
	    \vspace{-0.35cm}
	\end{subfigure} &
	\begin{subfigure}{.16\textwidth}
	    \centering
	    \includegraphics[width=1.0\linewidth]{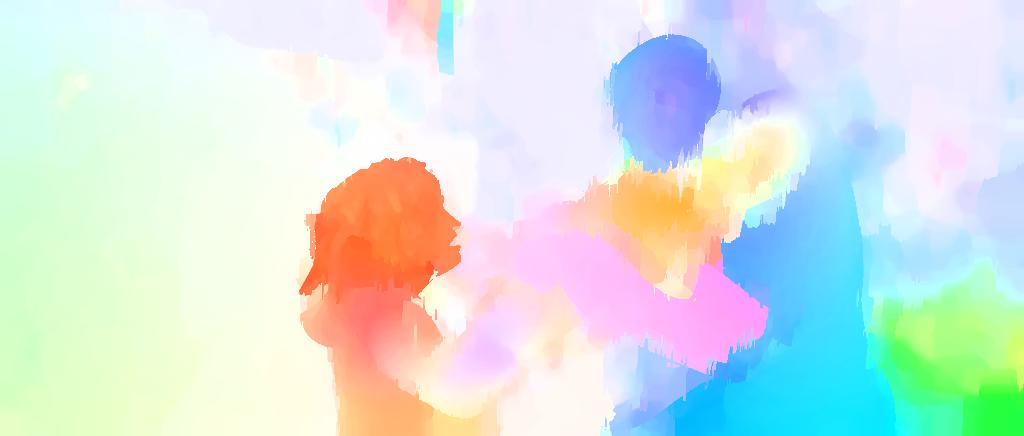}
	    \vspace{-0.35cm}
	\end{subfigure} &
	\begin{subfigure}{.16\textwidth}
	    \centering
	    \includegraphics[width=1.0\linewidth]{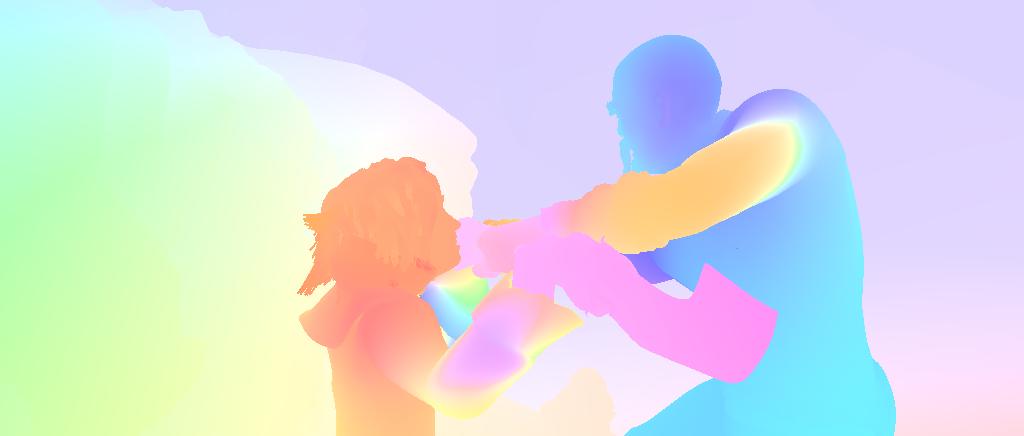}
	    \vspace{-0.35cm}
	\end{subfigure}
\\
    D & 	
    \begin{subfigure}{.16\textwidth}
	    \centering
	    \includegraphics[width=1.0\linewidth]{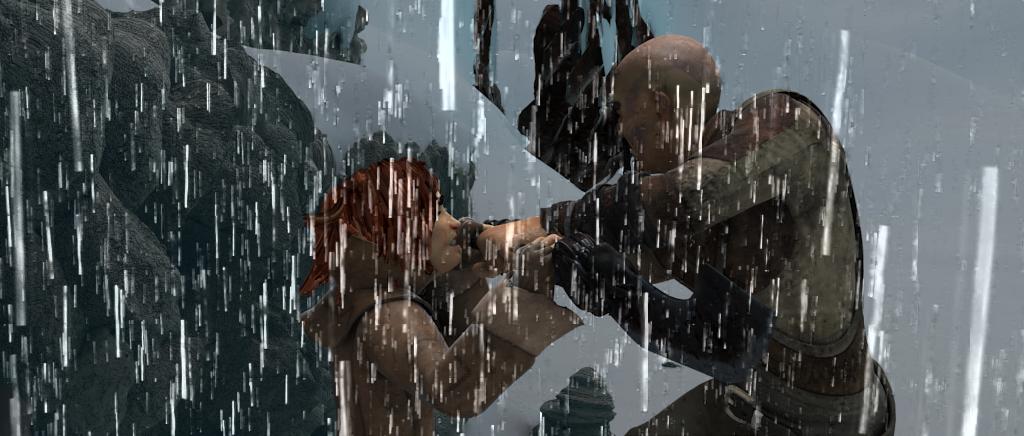}
	    \vspace{-0.35cm}
	\end{subfigure} &
    \begin{subfigure}{.16\textwidth}
	    \centering
	    \includegraphics[width=1.0\linewidth]{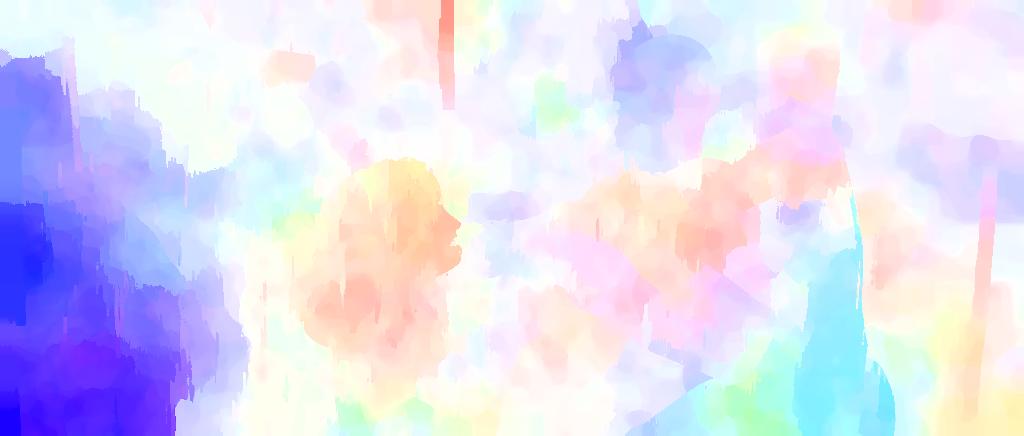}
	    \vspace{-0.35cm}
	\end{subfigure} &
    \begin{subfigure}{.16\textwidth}
	    \centering
	    \includegraphics[width=1.0\linewidth]{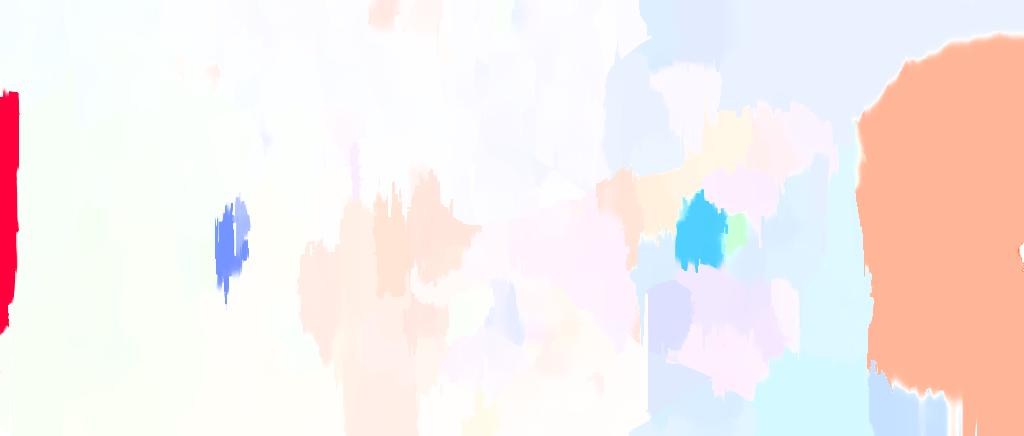}
	    \vspace{-0.35cm}
	\end{subfigure} &
    \begin{subfigure}{.16\textwidth}
	    \centering
	    \includegraphics[width=1.0\linewidth]{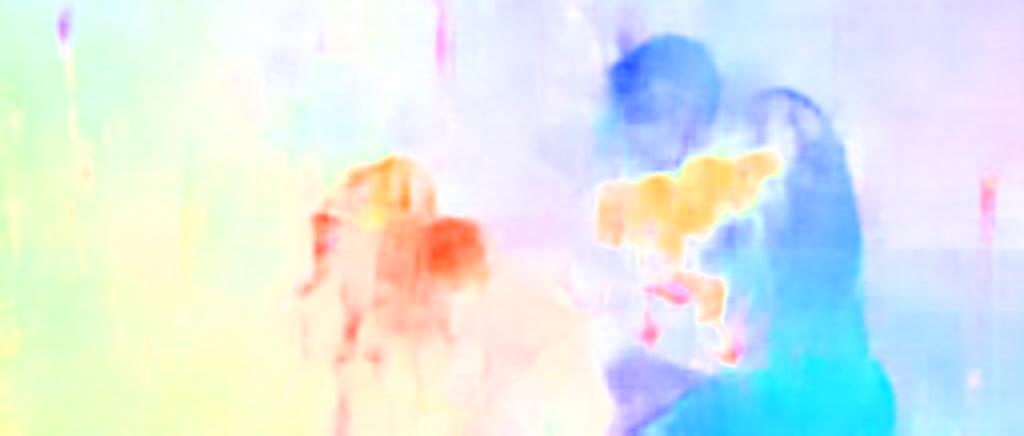}
	    \vspace{-0.35cm}
	\end{subfigure} &
	\begin{subfigure}{.16\textwidth}
	    \centering
	    \includegraphics[width=1.0\linewidth]{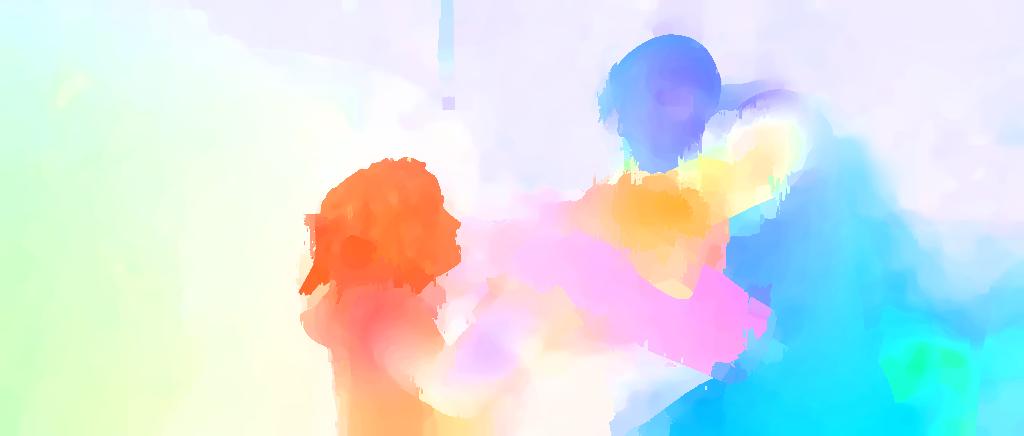}
	    \vspace{-0.35cm}
	\end{subfigure} &
	\begin{subfigure}{.16\textwidth}
	    \centering
	    \includegraphics[width=1.0\linewidth]{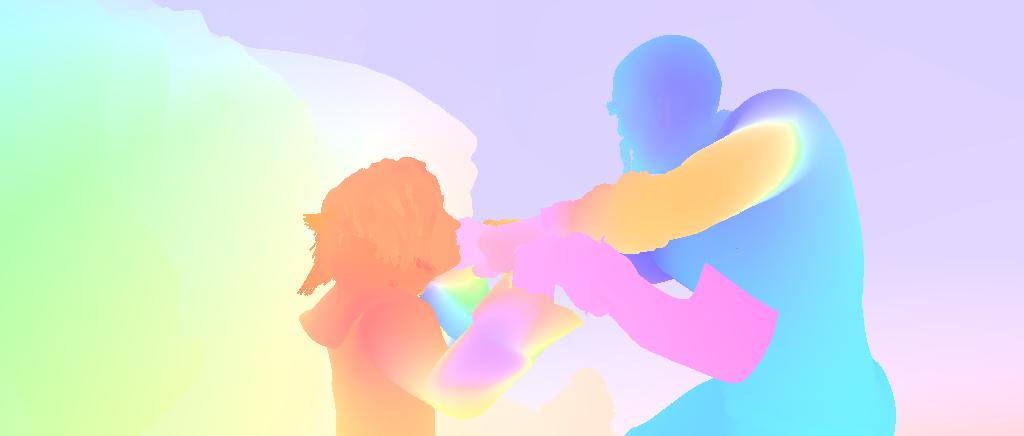}
	    \vspace{-0.35cm}
	\end{subfigure}
\\
   R & 	
    \begin{subfigure}{.16\textwidth}
	    \centering
	    \includegraphics[width=1.0\linewidth]{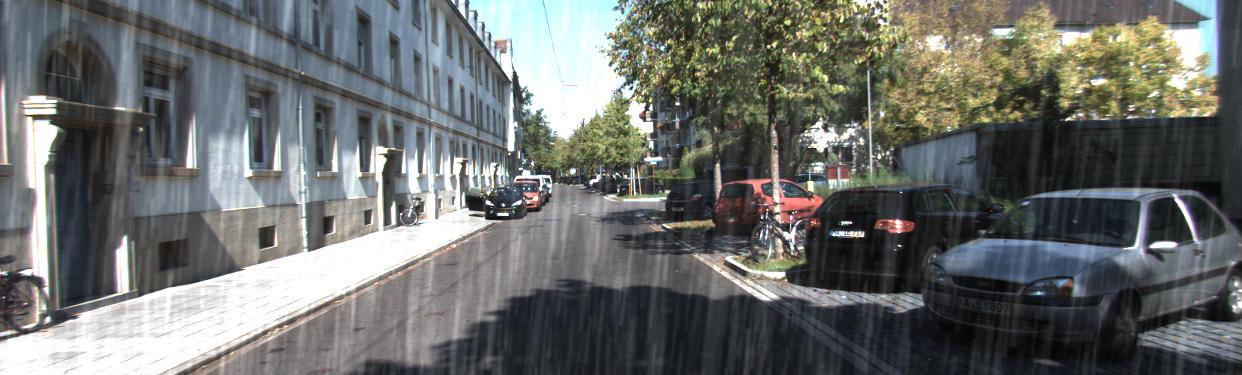}
	    \vspace{-0.35cm}
	\end{subfigure} &
    \begin{subfigure}{.16\textwidth}
	    \centering
	    \includegraphics[width=1.0\linewidth]{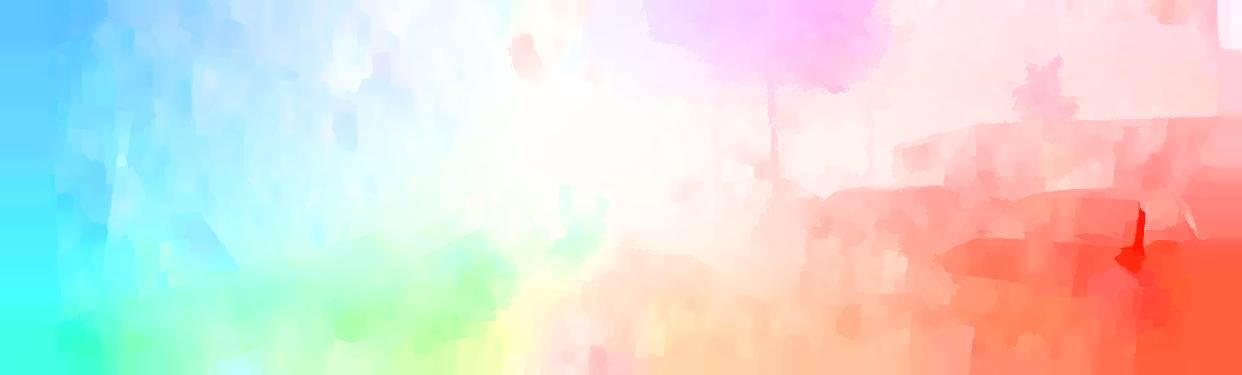}
	    \vspace{-0.35cm}
	\end{subfigure} &
    \begin{subfigure}{.16\textwidth}
	    \centering
	    \includegraphics[width=1.0\linewidth]{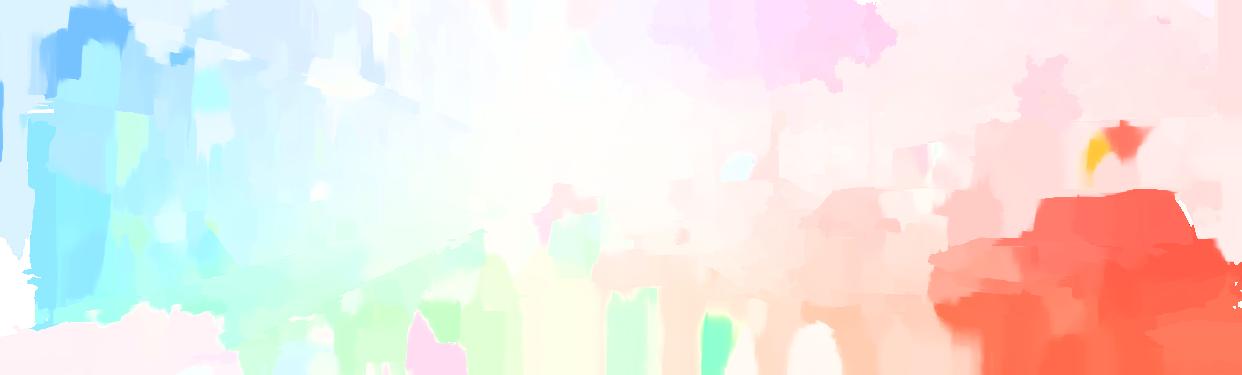}
	    \vspace{-0.35cm}
	\end{subfigure} &
    \begin{subfigure}{.16\textwidth}
	    \centering
	    \includegraphics[width=1.0\linewidth]{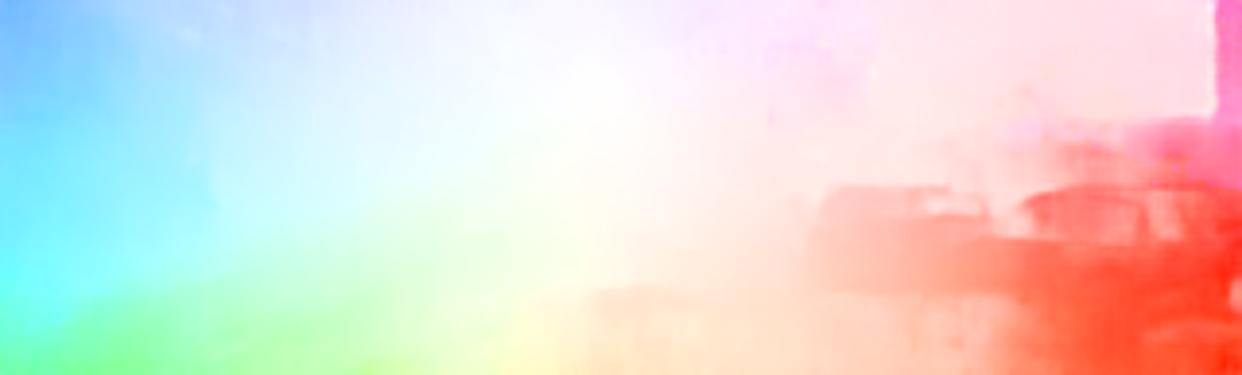}
	    \vspace{-0.35cm}
	\end{subfigure} &
	\begin{subfigure}{.16\textwidth}
	    \centering
	    \includegraphics[width=1.0\linewidth]{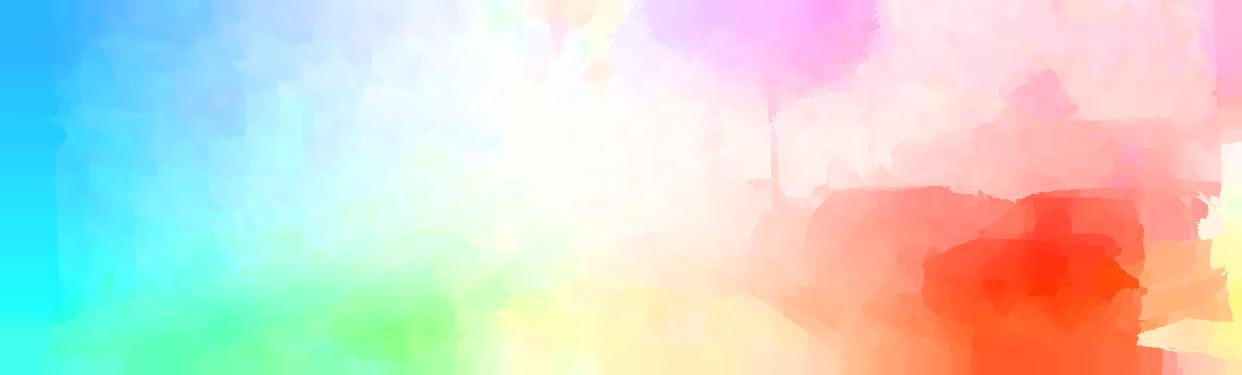}
	    \vspace{-0.35cm}
	\end{subfigure} &
	\begin{subfigure}{.16\textwidth}
	    \centering
	    \includegraphics[width=1.0\linewidth]{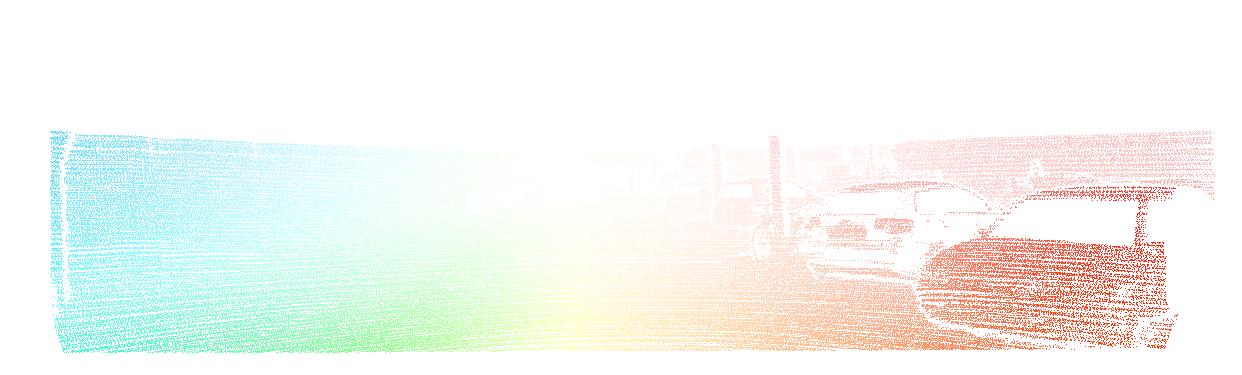}
	    \vspace{-0.35cm}
	\end{subfigure}
\\
    D & 	
    \begin{subfigure}{.16\textwidth}
	    \centering
	    \includegraphics[width=1.0\linewidth]{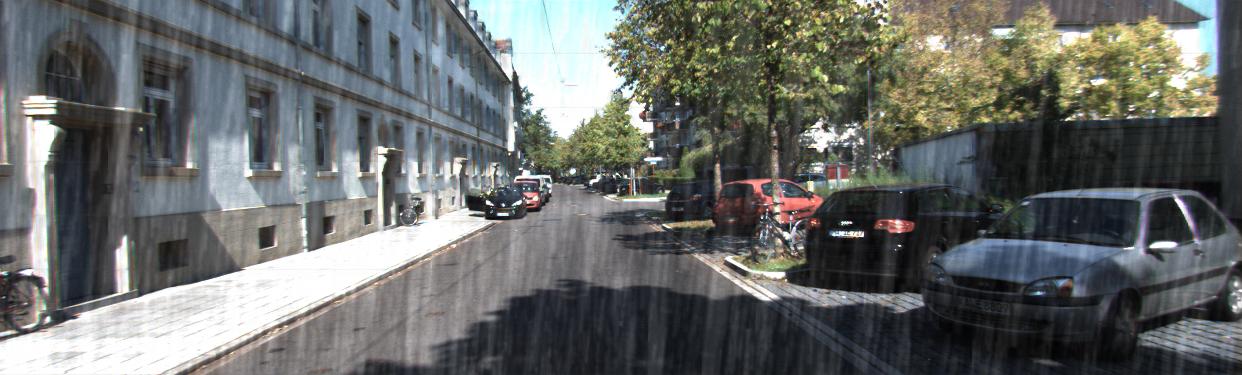}
	    \vspace{-0.35cm}
	\end{subfigure} &
    \begin{subfigure}{.16\textwidth}
	    \centering
	    \includegraphics[width=1.0\linewidth]{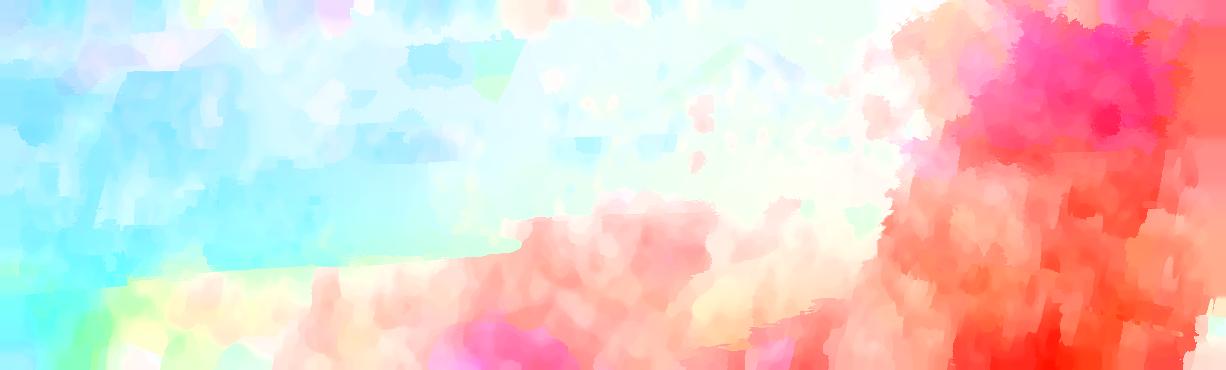}
	    \vspace{-0.35cm}
	\end{subfigure} &
    \begin{subfigure}{.16\textwidth}
	    \centering
	    \includegraphics[width=1.0\linewidth]{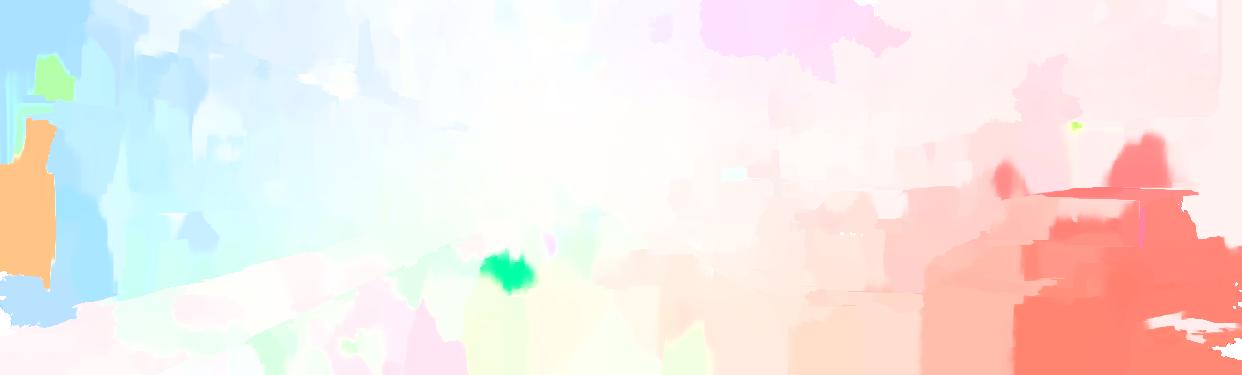}
	    \vspace{-0.35cm}
	\end{subfigure} &
    \begin{subfigure}{.16\textwidth}
	    \centering
	    \includegraphics[width=1.0\linewidth]{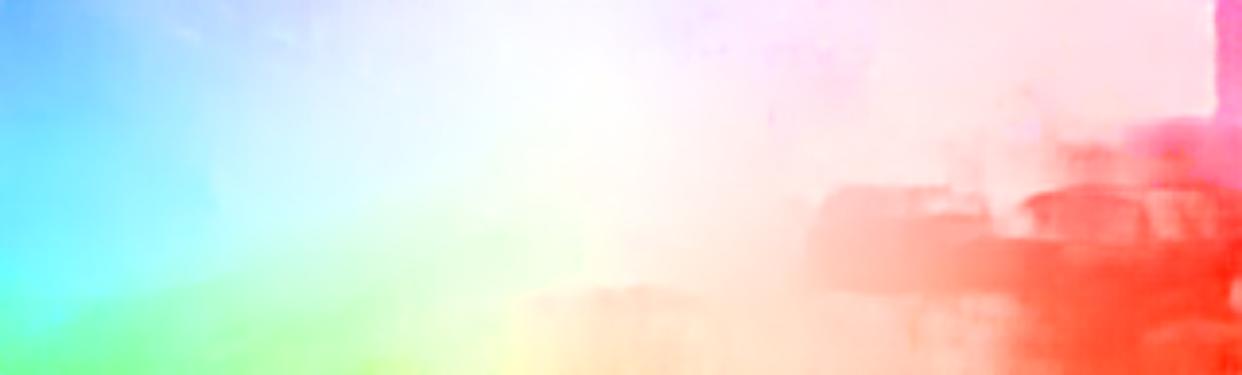}
	    \vspace{-0.35cm}
	\end{subfigure} &
	\begin{subfigure}{.16\textwidth}
	    \centering
	    \includegraphics[width=1.0\linewidth]{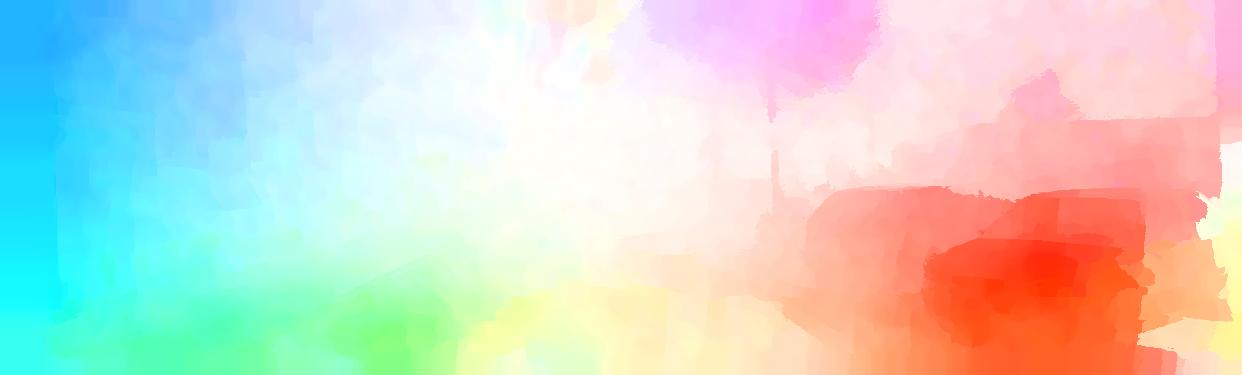}
	    \vspace{-0.35cm}
	\end{subfigure} &
	\begin{subfigure}{.16\textwidth}
	    \centering
	    \includegraphics[width=1.0\linewidth]{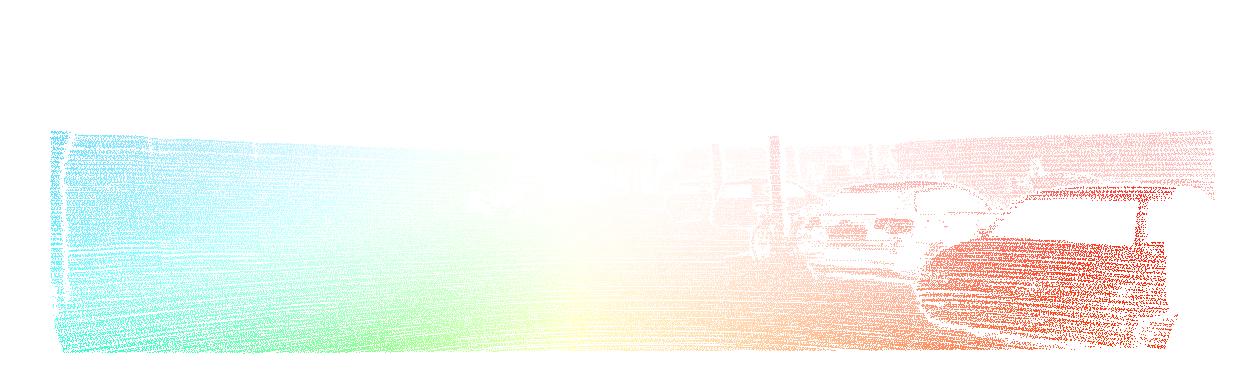}
	    \vspace{-0.35cm}
	\end{subfigure}
\\
\end{tabular}
\caption{Method comparison on Middlebury, MPI Sintel, and KITTI datasets, all rendered with rain. The first column "R" and "D" represent synthesized rain sequences and the same sequences after \cite{YangTFLGY16}'s de-rain method. (\textbf{Best zoom in on screen}).}
\label{fig:SyntheticResult}
\end{figure*}

\subsection{Flying Vehicles with Rain}
To provide a real rain dataset that can quantitatively evaluate methods, we carefully capture two consecutive frames of a static background in a rainy scene and manually crop out all the moving objects (if any) so that every object except the rain is static (shown in Fig.~\ref{fig:StaticAnalysis}). We cut each of the resultant image frames into 12 corresponding patches as a background, each of which has a resolution of 512$\times$384. For the foreground objects, we collect 3D vehicle models from the 'car' category in Trimble 3D Warehouse\footnote{General model licence of Trimble Inc.} and project them onto 24 2D images each with different poses. In order to render the rain onto the vehicles, we extract the rain streaks and fog-like rain accumulation from the captured rain frames using \cite{YangTFLGY16} and \cite{DBLP:journals/tip/CaiXJQT16} respectively, and apply the rain effects extracted from the regions, which are covered by the vehicles, back on top of those vehicles.

To generate motion, we randomly sample 2D Euclidean transformation parameters from a family of Gaussian distribution for the background and the vehicle models. The background motion contains horizontal and vertical translation parameters (\(T_x^{bg}, T_y^{bg}\)) sampled from a Gaussian distribution $ X \sim \mathcal{N}(0, 10).$  The background motion can be interpreted as camera translation. The foreground car models contain translation parameters (\(T_x^{fg}, T_y^{fg}\)) and rotation parameter \(\theta^{fg}\) which are sample from $X \sim \mathcal{N}(0,5)$. The car models are uniformly distributed in the first image and are rendered on the second image using the transformation parameters. As a result, this dataset contains 12 real rain image pairs with optical flow ground truth and is meant for algorithm testing and evaluation\footnote {This dataset is available at \url{https://github.com/liruoteng/FlyingVehiclesWithRain}}. Because the vehicles are randomly generated in the rainy scene, we call it "Flying Vehicles with Rain" dataset.

\setlength{\tabcolsep}{2pt}
\begin{table}
\resizebox{0.49\textwidth}{!}{%
\begin{tabular}{c|cc|cc|cc|cc}
\hline
\hline
Method                          & \multicolumn{2}{c|}{Middlebury} & \multicolumn{2}{c|}{Sintel}  & \multicolumn{2}{c|}{KITTI2012} &  \multicolumn{2}{c}{FVR} \\
{}                              & Rain          & De-rain         & Rain         & De-rain       & Rain       & De-rain       & Rain   & De-rain  \\
\hline
Classic+NL \cite{Sun:CVPR:10}   & 0.90          & 0.60           & 12.89        & 9.51           & 9.17       & 9.14          & 2.79   & 3.20   \\
\hline
LDOF \cite{LDOF}                & 0.90          & 0.66           & 18.26        & 11.84          & 10.17      & 9.90          & 3.83   & 4.09  \\
\hline
SP-MBP \cite{Yuli2015SPMBP}     & 0.93          & 0.60           & 14.18        & 7.33           & 15.71      & 15.94         & 5.18   & 5.94  \\
\hline
FlowNetS \cite{DFIB15}          & 2.58          & 1.54           & 50.90        & 21.96          & 17.43      & 18.73         & 4.75   & 5.05 \\
\hline
FlowNetS-Rain                   & 1.45          & 1.42           & 7.91         & 6.21           & 6.84       & 6.91          & 2.25   & 2.31 \\
\hline
\hline
Ours                            & \textbf{0.30} & \textbf{0.332}  & \textbf{6.11}&\textbf{5.27}  &\textbf{6.65} & \textbf{6.67} & \textbf{1.90}  & \textbf{2.15} \\
\hline
\hline
\end{tabular}}
\caption{A comparison of our algorithm with several top-performing methods on synthesized rain datasets. 'De-rain' indicates the results of each method performed on the sequences after Yang et al.'s \cite{YangTFLGY16} de-rain method.}
\label{table:quantitative}
\end{table}

\subsection{Synthetic Rain Results}
In the synthetic rain experiment, we compare our algorithm with some classic conventional methods, i.e. Classic+NL \cite{Sun:CVPR:10}, LDOF \cite{LDOF}, and SP-MBP \cite{Yuli2015SPMBP}, as well as deep learning method FlowNet \cite{DFIB15} on the three existing datasets rendered with rain streaks. For a fairer comparison, we also utilize the recent deraining method \cite{YangTFLGY16} as preprocessing and compare the results obtained in this way. The quantitative results are shown in Table~\ref{table:quantitative}. Fig.~\ref{fig:SyntheticResult} shows the qualitative results of these comparisons. The original synthesized rain data examples are denoted with 'R' in the beginning of every row. Those examples after deraining preprocessing are denoted with 'D'. We have tuned the parameters of method \cite{Sun:CVPR:10} \cite{LDOF} to show their best performance in the Table~\ref{table:quantitative}. Due to the limited space, we include the quantitative and qualitative results of our method performed on the aforementioned three existing \textbf{clean} datasets (\ie no rain) in the supplementary material. Though our method discards some fine texture details and relies more on the structure information, it still performs well on the clean datasets.

Since the training data of FlowNetS \cite{DFIB15} does not include any rain, the FlowNetS model may not perform well under rain conditions. Hence, we render the Flying Chair dataset \cite{DFIB15} with synthetic rain streaks using the same rain streak model as the test dataset (shown in Fig.~\ref{fig:SyntheticResult}) following \cite{Garg:2006}. We then train the FlowNet network using the same parameter as FlowNetS provided by the authors on this Flying Chair dataset with rain. The model is denoted as FlowNetS-rain.

\begin{figure*}
\centering
\setlength{\tabcolsep}{1.0pt}
\begin{tabular}{@{} cccccc @{}}
  First frame & Classic + NL\cite{Sun:CVPR:10} & SPM-BP \cite{Yuli2015SPMBP} & FlowNetS-rain \cite{DFIB15} & Ours  & Ground truth \\
    \begin{subfigure}{.16\textwidth}
	    \centering
	    \includegraphics[width=1.0\linewidth]{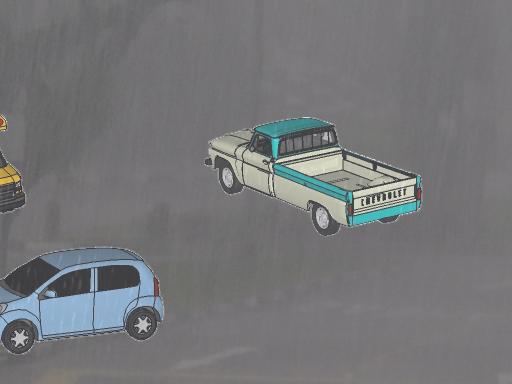}
	    \vspace{-0.35cm}
	\end{subfigure} &
    \begin{subfigure}{.16\textwidth}
	    \centering
	    \includegraphics[width=1.0\linewidth]{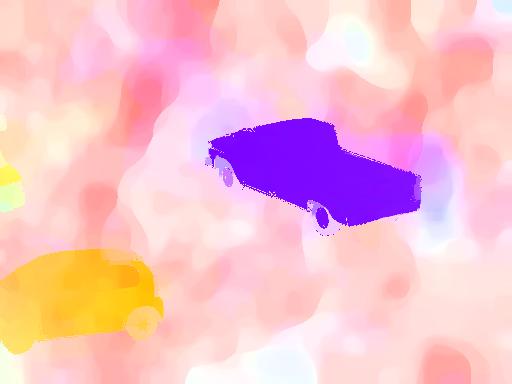}
	    \vspace{-0.35cm}
	\end{subfigure} &
    \begin{subfigure}{.16\textwidth}
	    \centering
	    \includegraphics[width=1.0\linewidth]{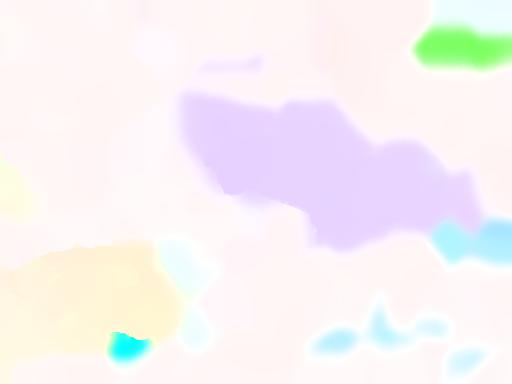}
	    \vspace{-0.35cm}
	\end{subfigure} &
    \begin{subfigure}{.16\textwidth}
	    \centering
	    \includegraphics[width=1.0\linewidth]{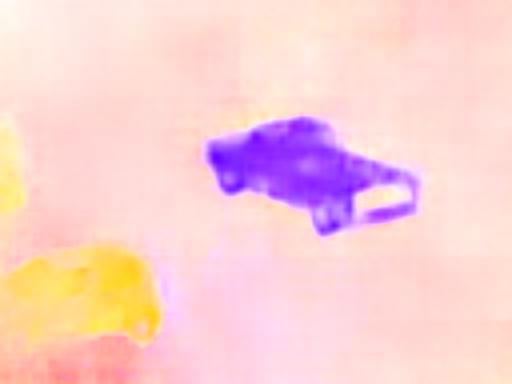}
	    \vspace{-0.35cm}
	\end{subfigure} &
	\begin{subfigure}{.16\textwidth}
	    \centering
	    \includegraphics[width=1.0\linewidth]{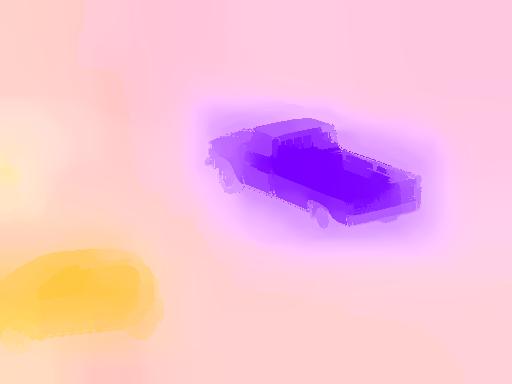}
	    \vspace{-0.35cm}
	\end{subfigure} &
	\begin{subfigure}{.16\textwidth}
	    \centering
	    \includegraphics[width=1.0\linewidth]{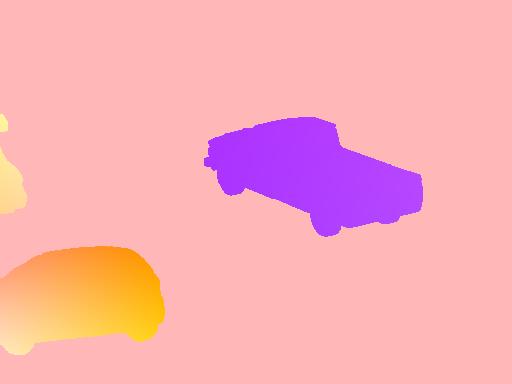}
	    \vspace{-0.35cm}
	\end{subfigure}
\\	
    \begin{subfigure}{.16\textwidth}
	    \centering
	    \includegraphics[width=1.0\linewidth]{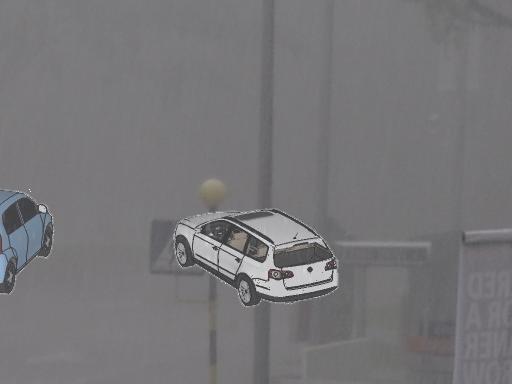}
	    \vspace{-0.35cm}
	\end{subfigure} &
    \begin{subfigure}{.16\textwidth}
	    \centering
	    \includegraphics[width=1.0\linewidth]{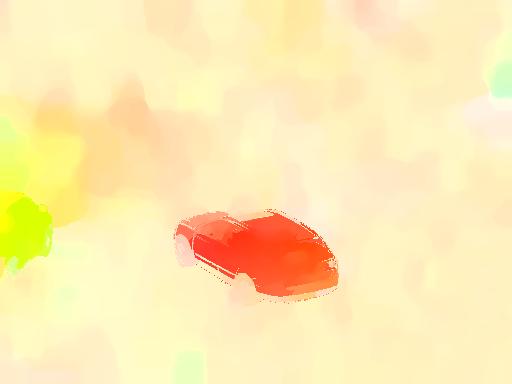}
	    \vspace{-0.35cm}
	\end{subfigure} &
    \begin{subfigure}{.16\textwidth}
	    \centering
	    \includegraphics[width=1.0\linewidth]{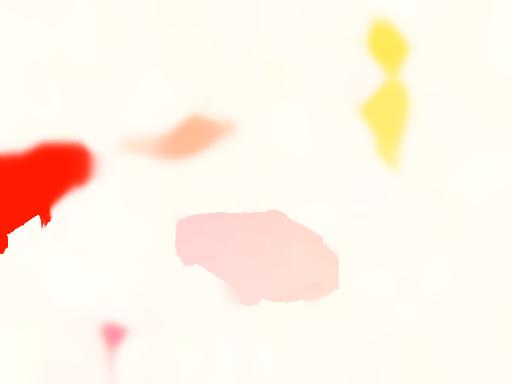}
	    \vspace{-0.35cm}
	\end{subfigure} &
    \begin{subfigure}{.16\textwidth}
	    \centering
	    \includegraphics[width=1.0\linewidth]{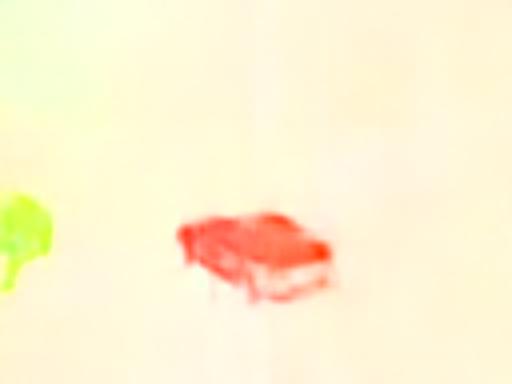}
	    \vspace{-0.35cm}
	\end{subfigure} &
	\begin{subfigure}{.16\textwidth}
	    \centering
	    \includegraphics[width=1.0\linewidth]{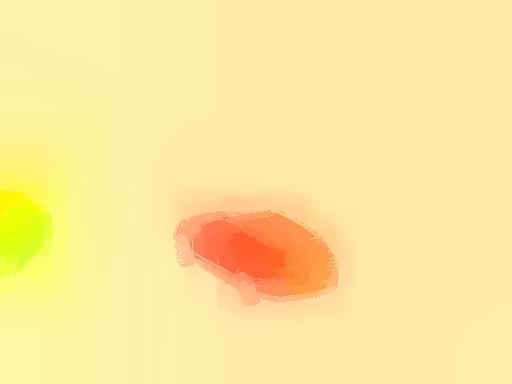}
	    \vspace{-0.35cm}
	\end{subfigure} &
	\begin{subfigure}{.16\textwidth}
	    \centering
	    \includegraphics[width=1.0\linewidth]{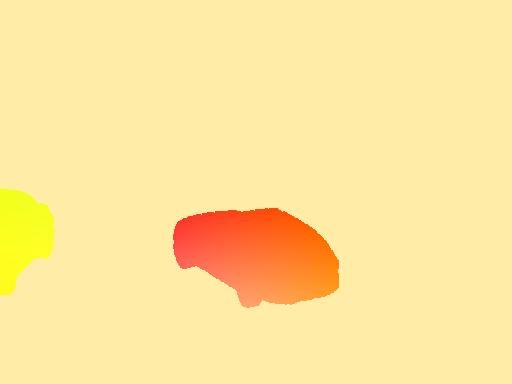}
	    \vspace{-0.35cm}
	\end{subfigure}
\\
\end{tabular}
\caption{ Method comparison on Flying Vehicle with Rain (FVR) dataset. (\textbf{Best viewed on screen}).}
\label{fig:FVRResult}
\end{figure*}
\begin{figure*}
\centering
\setlength{\tabcolsep}{1.0pt}
\begin{tabular}{@{} ccccccc @{}}
 & Input Image & Classic & SPMBP & FlowNetS & FlowNetS-rain & Ours  \\
    (a) & 	
    \begin{subfigure}{.16\textwidth}
	    \centering
	    \includegraphics[width=1.0\linewidth]{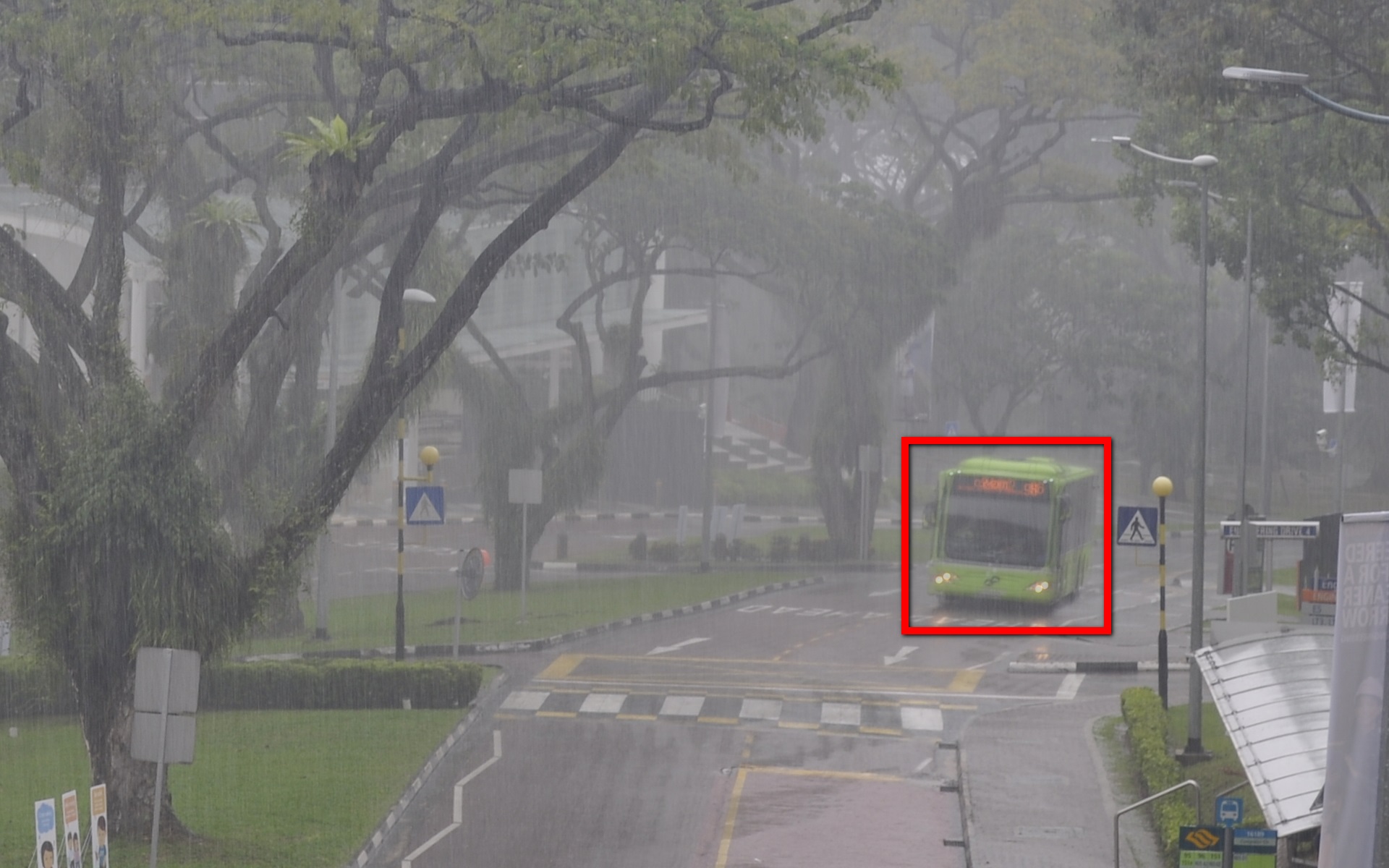}
	    \vspace{-0.35cm}
	\end{subfigure} &
	\begin{subfigure}{.16\textwidth}
	    \centering
	    \includegraphics[width=1.0\linewidth]{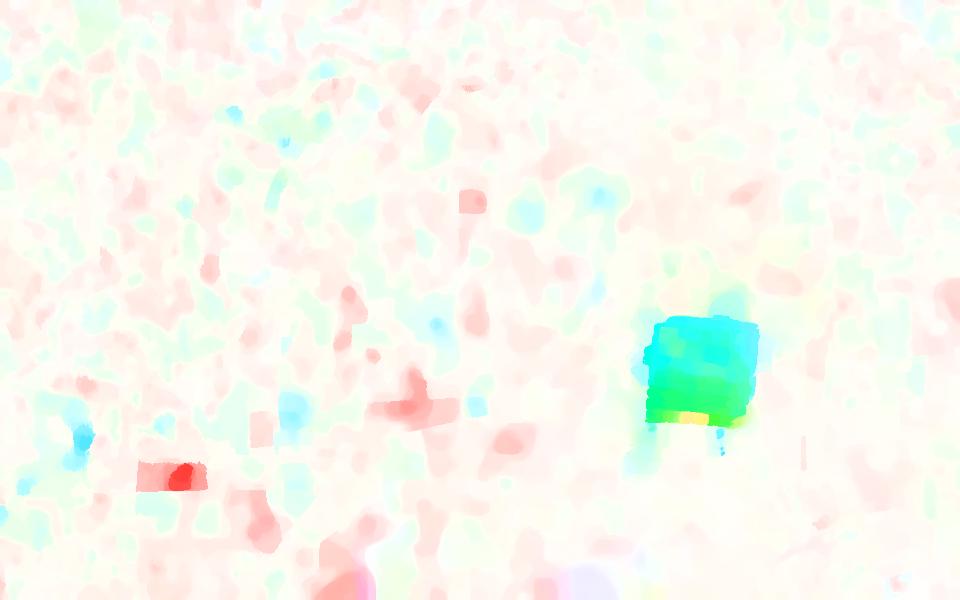}
	    \vspace{-0.35cm}
	\end{subfigure} &
    \begin{subfigure}{.16\textwidth}
	    \centering
	    \includegraphics[width=1.0\linewidth]{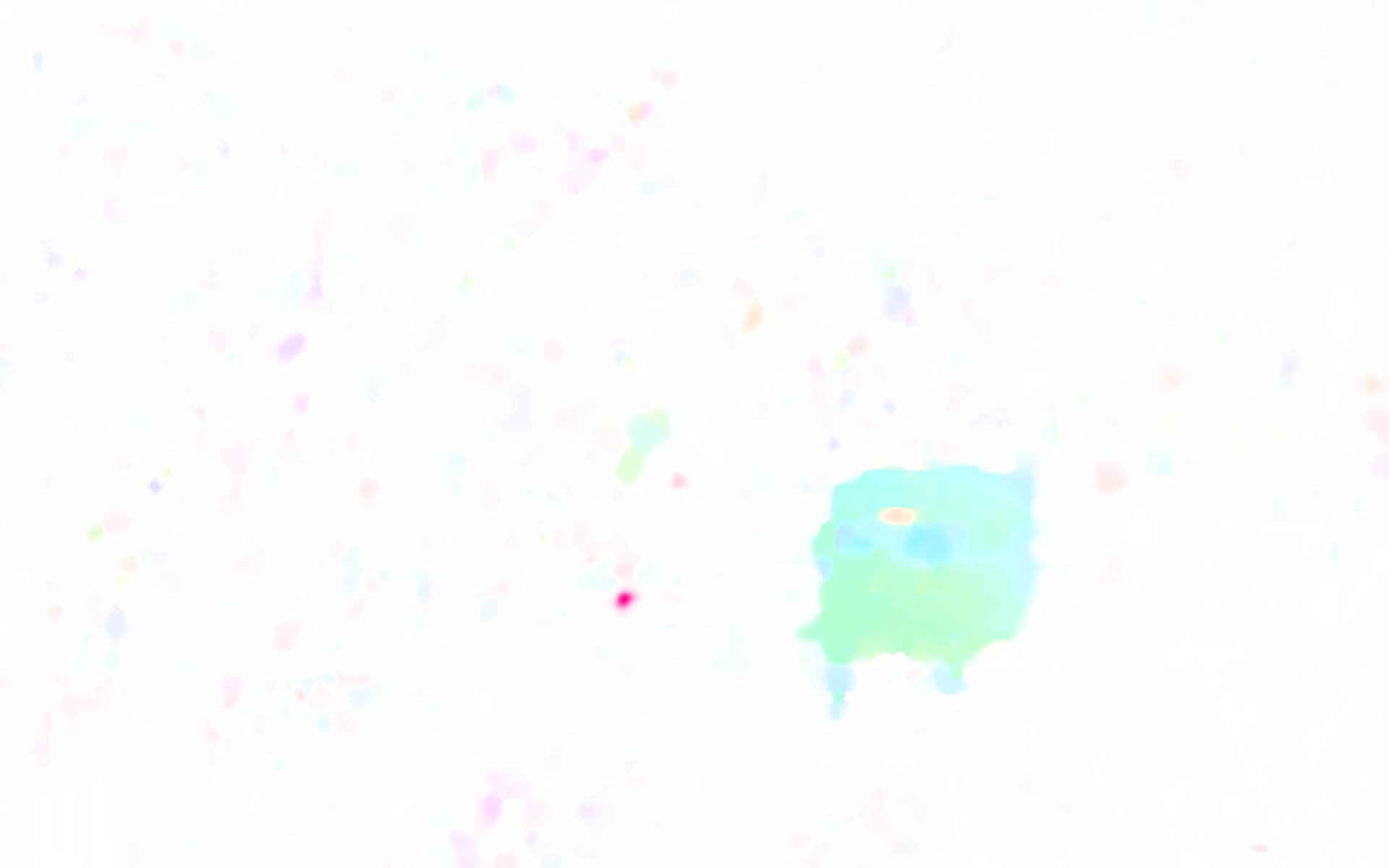}
	    \vspace{-0.35cm}
	\end{subfigure} &
    \begin{subfigure}{.16\textwidth}
	    \centering
	    \includegraphics[width=1.0\linewidth]{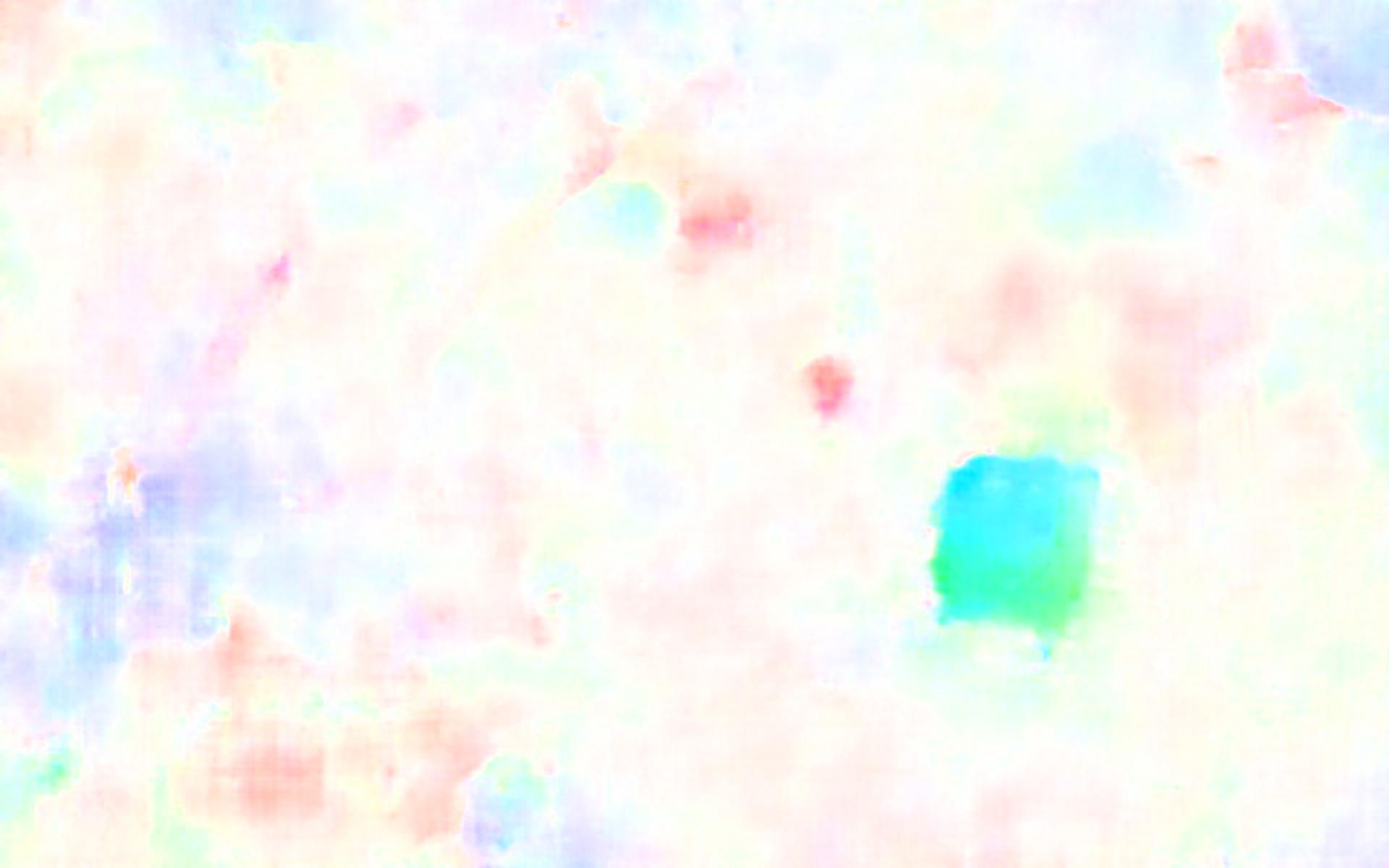}
	    \vspace{-0.35cm}
	\end{subfigure} &
    \begin{subfigure}{.16\textwidth}
	    \centering
	    \includegraphics[width=1.0\linewidth]{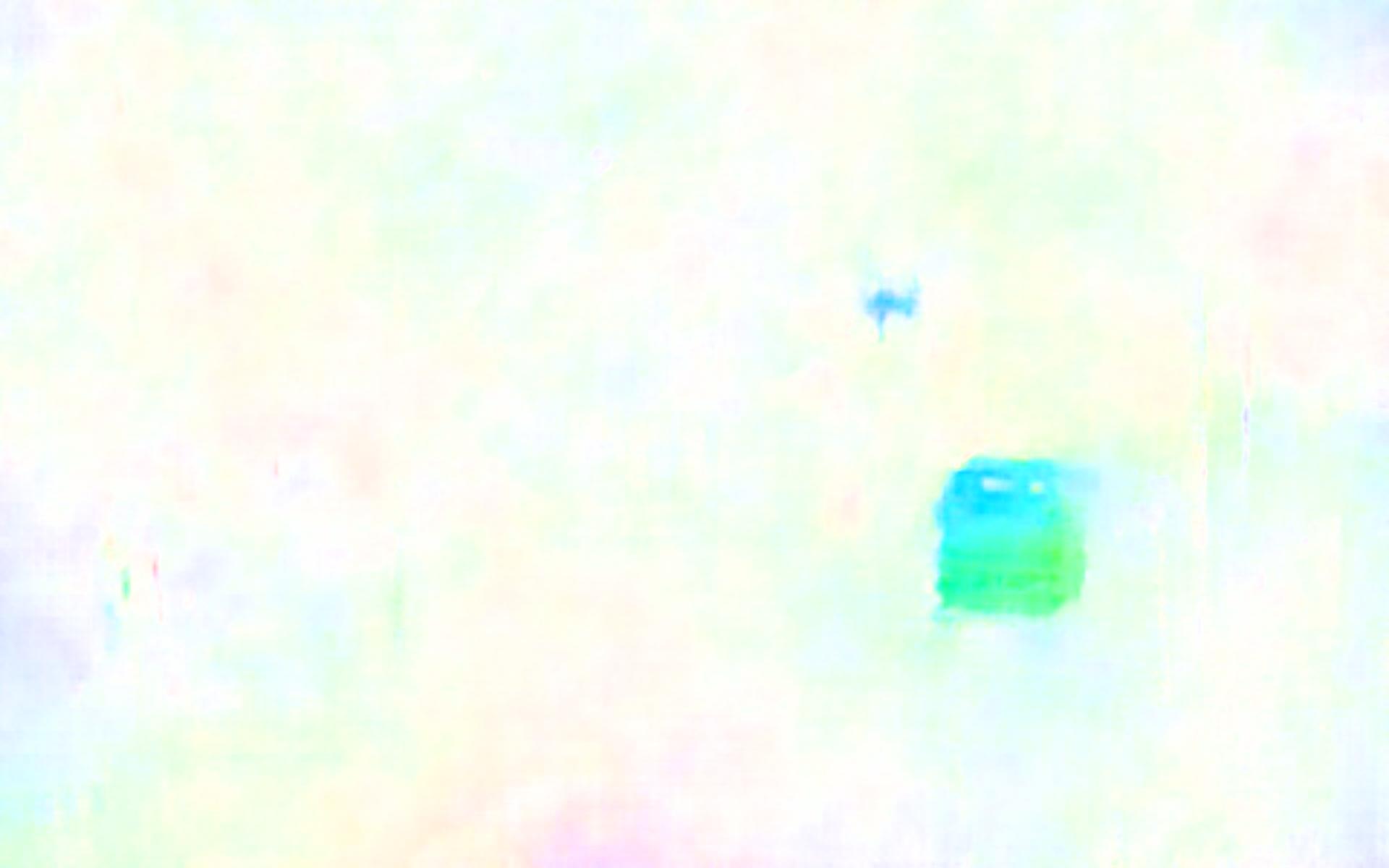}
	    \vspace{-0.35cm}
	\end{subfigure} &
	\begin{subfigure}{.16\textwidth}
	    \centering
	    \includegraphics[width=1.0\linewidth]{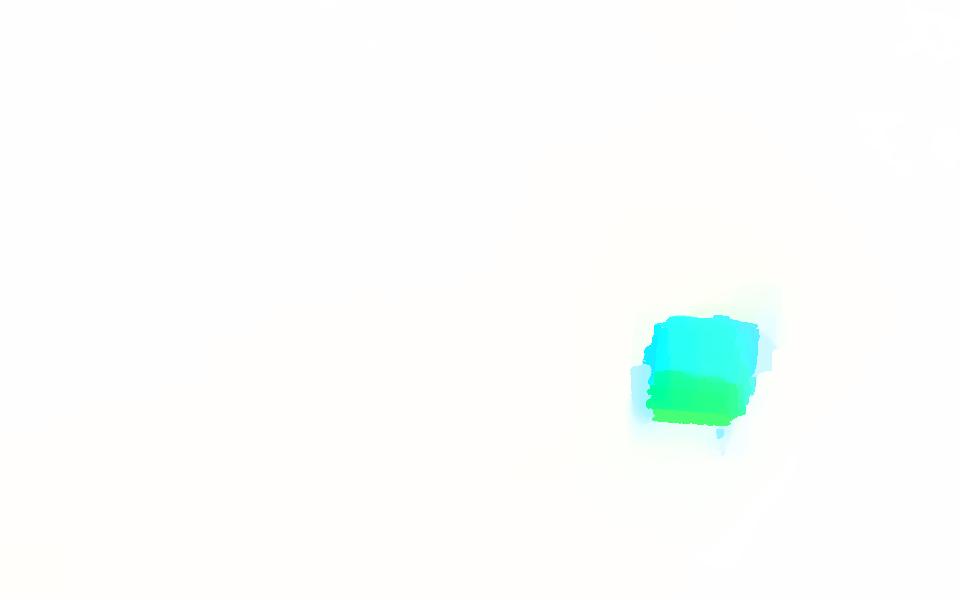}
	    \vspace{-0.35cm}
	\end{subfigure}
\\
    (b) & 	
    \begin{subfigure}{.16\textwidth}
	    \centering
	    \includegraphics[width=1.0\linewidth]{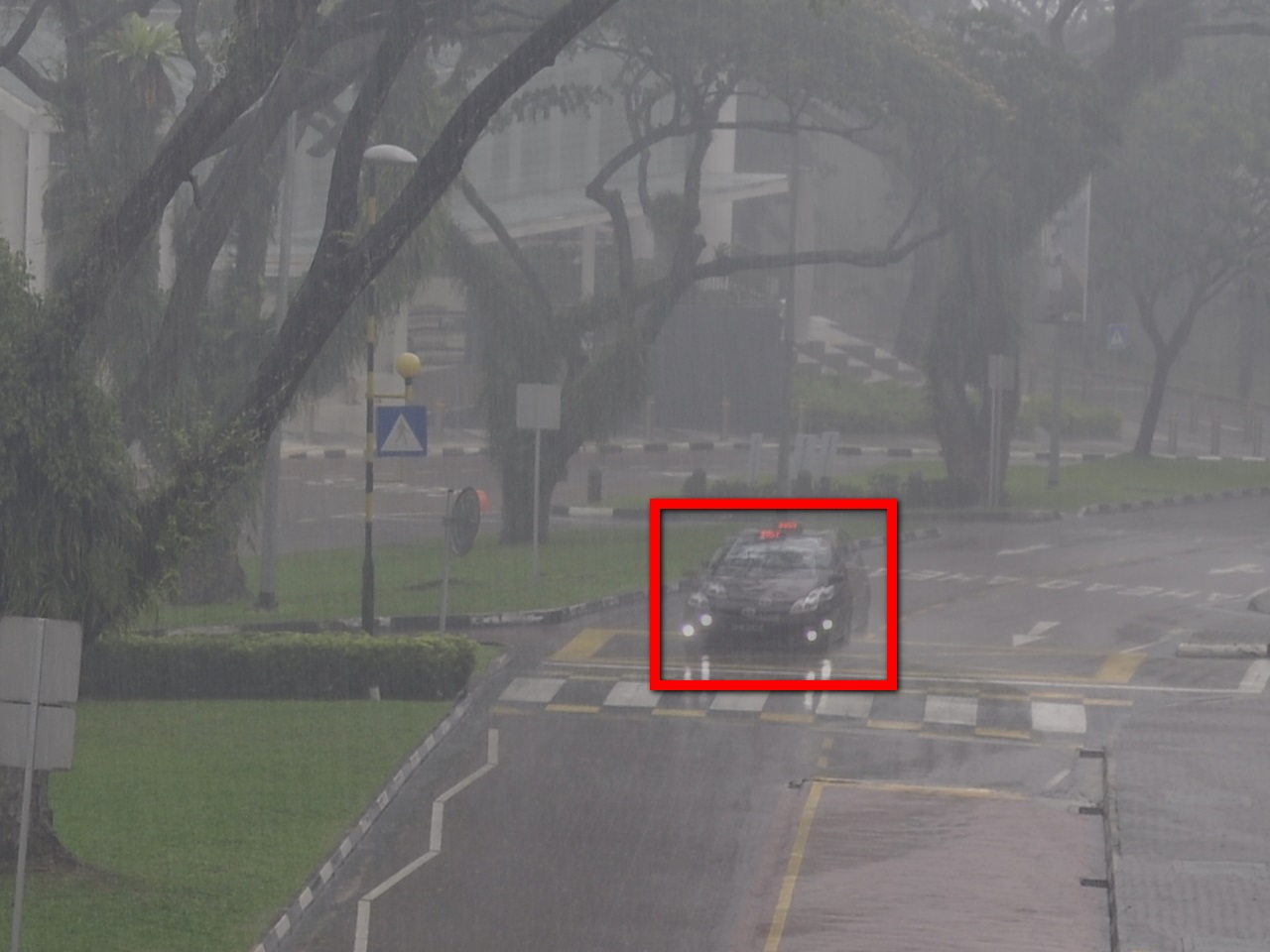}
	    \vspace{-0.35cm}
	\end{subfigure} &
	\begin{subfigure}{.16\textwidth}
	    \centering
	    \includegraphics[width=1.0\linewidth]{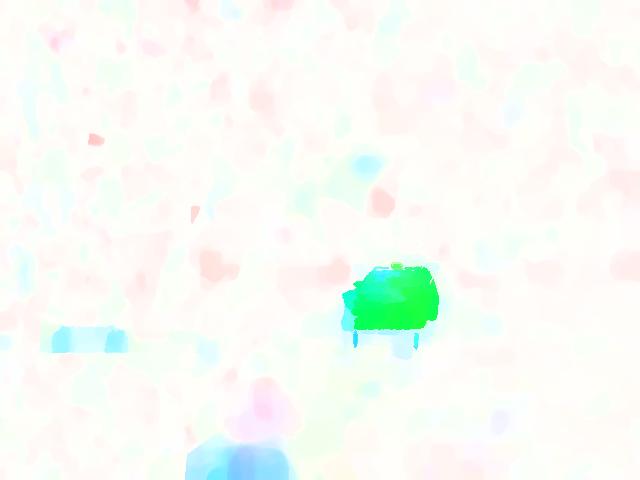}
	    \vspace{-0.35cm}
	\end{subfigure} &
    \begin{subfigure}{.16\textwidth}
	    \centering
	    \includegraphics[width=1.0\linewidth]{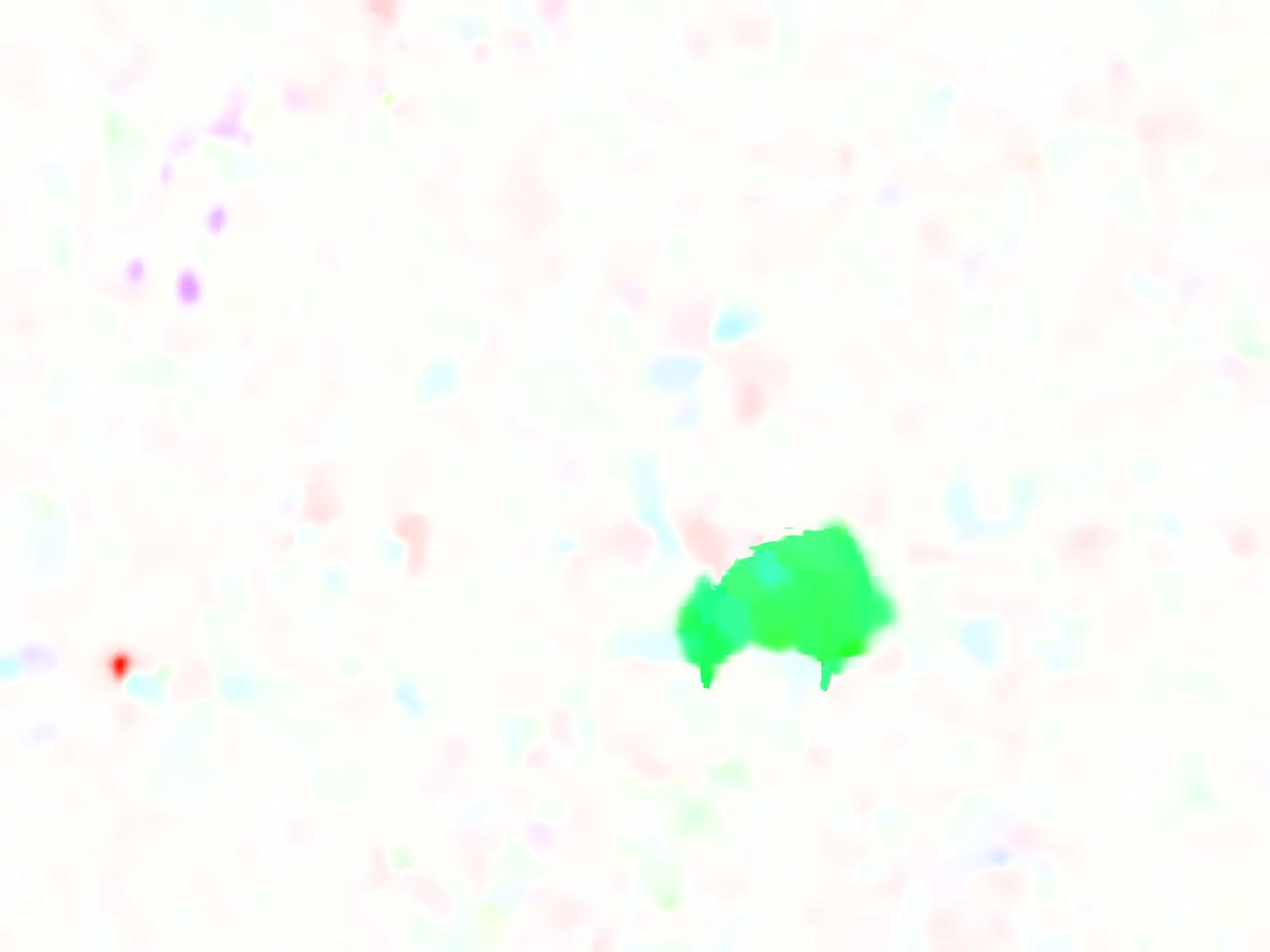}
	    \vspace{-0.35cm}
	\end{subfigure} &
    \begin{subfigure}{.16\textwidth}
	    \centering
	    \includegraphics[width=1.0\linewidth]{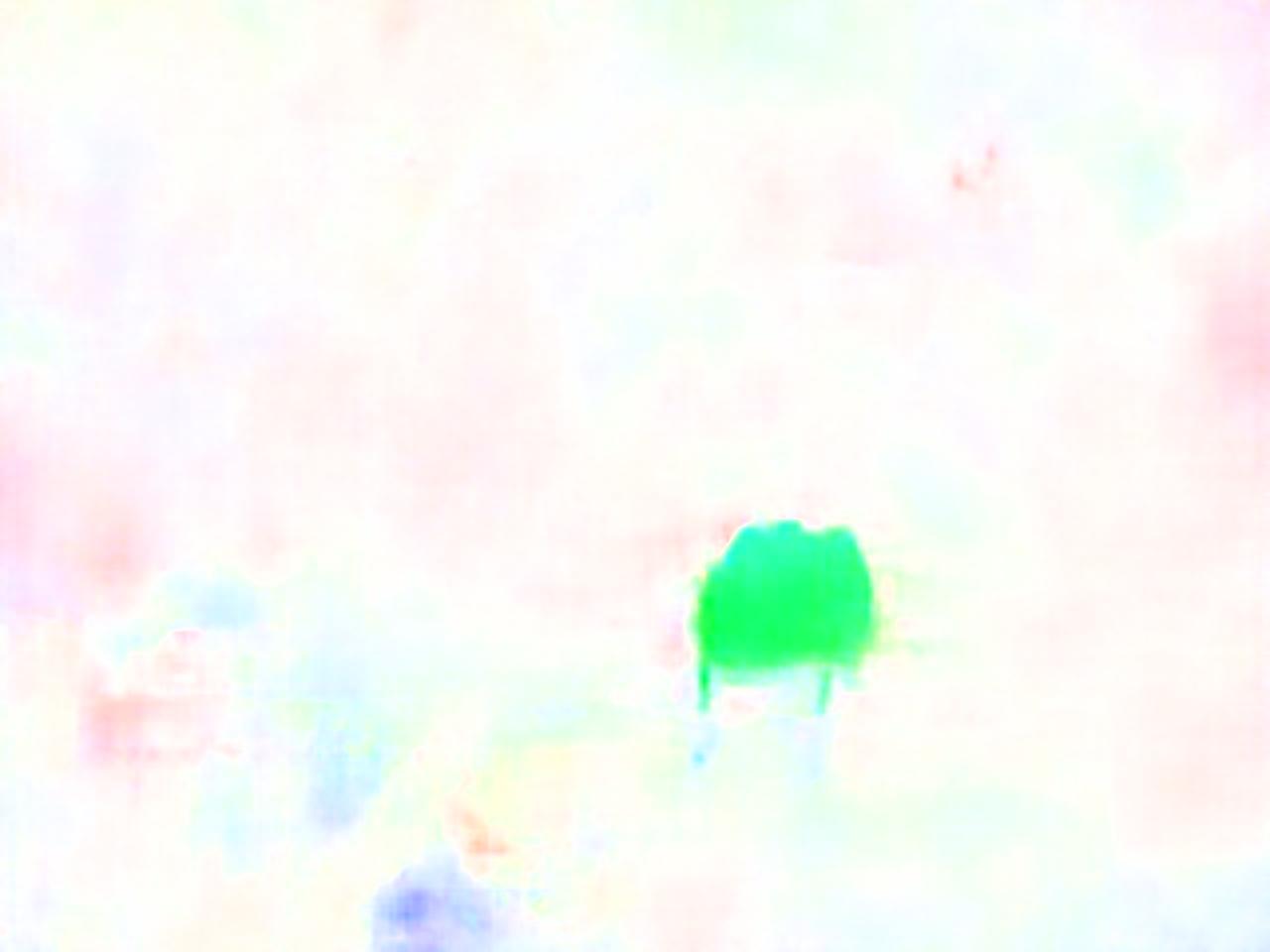}
	    \vspace{-0.35cm}
	\end{subfigure} &
    \begin{subfigure}{.16\textwidth}
	    \centering
	    \includegraphics[width=1.0\linewidth]{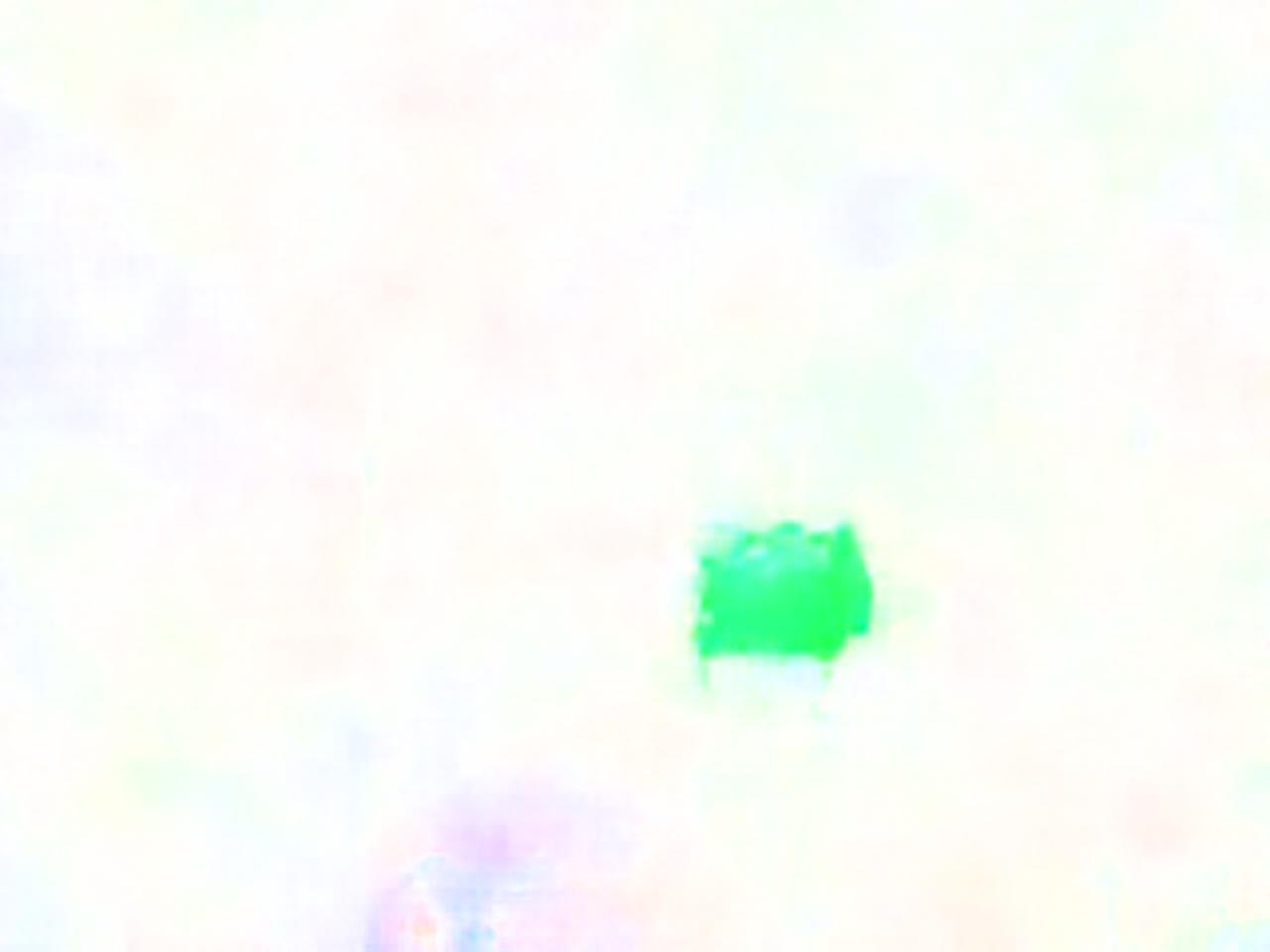}
	    \vspace{-0.35cm}
	\end{subfigure} &
	\begin{subfigure}{.16\textwidth}
	    \centering
	    \includegraphics[width=1.0\linewidth]{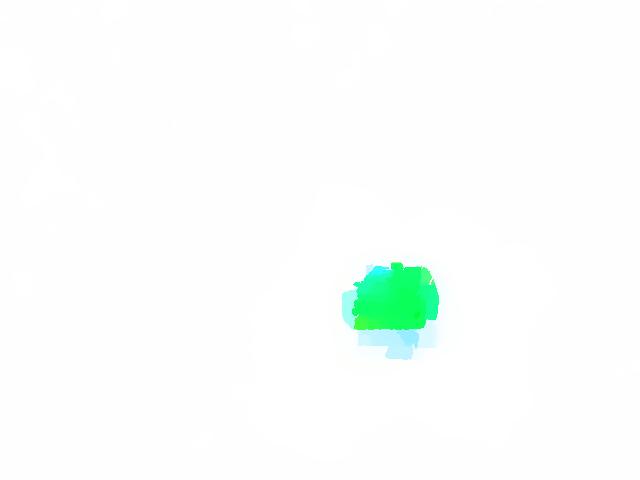}
	    \vspace{-0.35cm}
	\end{subfigure}
\\
\end{tabular}
\caption{Method comparison on real rainy scenes with different severity level. The red box indicates the only motion in each image pair.  (\textbf{Best zoom in on screen}).}
\label{fig:realrainqualitative}
\end{figure*}

\subsection{Real Rain Results}
For evaluations in real rain scenarios, we first present analysis of a basic static scene, followed by qualitative and quantitative results of real rain sequences with moving cameras. Due to the lack of ground-truths for real rain scenes, we compare different methods on the FVR dataset quantitatively, because it contains real rain streaks and rain accumulation effect on a real background scene.

\paragraph{Static Scene Analysis} To verify the correctness and effectiveness of our algorithm, we perform a sanity check on the baseline algorithms and our algorithm on the static real-rain image pairs at one quarter (649$\times$362) of the original image size due to the GPU memory constraint when testing FlowNetS\cite{DFIB15}. Since this is a static scene under heavy rain, the true optical flow for the background should be zero everywhere. From Fig.~\ref{fig:CoverPage} one can see that the baseline methods produce erroneous flow due to the motion of the rain. In comparison, the result of our algorithm is shown in the bottom row and last column of Fig.\ref{fig:StaticAnalysis}. The average magnitude of our flow field is \textbf{0.000195} pixel, which is essentially zero flow. Fig.~\ref{fig:StaticAnalysis} also shows the intermediate estimated flow fields of our algorithm. One can observe that the flow field is being increasingly cleaned up in each iteration.

\paragraph{Quantitative Result} We compare the baseline methods with our algorithm on the FVR dataset for quantitative evaluation as shown in the last column of Table~\ref{table:quantitative}. We also apply the de-rain pre-processing using \cite{YangTFLGY16} on the FVR dataset and compare the results of ours with the baseline methods. The results show that our algorithm outperforms those of the baseline methods even with the derain preprocessing applied. Some qualitative examples are shown in Fig.\ref{fig:FVRResult}.

\paragraph{Qualitative Result} Finally, we compare our algorithm with the aforementioned methods on real rainy sequences. The qualitative results are shown in Fig.~\ref{fig:realrainqualitative}. In order to better visualize the object motion, we select the rain sequences with only one object motion included in the red box in the images (See more results in the supplementary material).
One can see that the classic and SPMBP methods generate sporadic erroneous flow on the background regions due to the rain. The result of FlowNetS-rain has a clearer background flow field than FlowNetS, which benefits from the synthesized rain training data. However, one can still observe the 'colorful' background flow and the blurry motion boundary. Our algorithm is able to generate clean blank background without losing the sharp motion boundaries of the moving car.

\section{Conclusion}
We have introduced a robust algorithm for optical flow estimation in rainy scenes. To come to grips with rain streaks and rain accumulation effect, we propose the residue channel and a layer decomposition scheme. In our experiments, the quantitative and qualitative results show that our method outperforms other state-of-the-art methods on both synthetic and real rain datasets.

{\small
\bibliographystyle{ieee}
\bibliography{egbib}
}

\end{document}